\documentclass{article}
\usepackage[dvipsnames]{xcolor}
\definecolor{Gray}{gray}{0.9}
\usepackage{microtype}
\usepackage{graphicx}
\usepackage{subfigure}
\usepackage{booktabs} 

\usepackage{hyperref}

\usepackage[accepted]{icml2024}

\usepackage{amsmath}
\usepackage{amssymb}
\usepackage{mathtools}
\usepackage{amsthm}

\usepackage[capitalize,noabbrev]{cleveref}

\theoremstyle{plain}

\theoremstyle{definition}

\theoremstyle{remark}

\usepackage[textsize=tiny]{todonotes}

\usepackage{etoc}
\usepackage{framed} 

\usepackage{etoc}
\etocdepthtag.toc{mtchapter}
\etocsettagdepth{mtchapter}{subsection}
\etocsettagdepth{mtappendix}{none}

\usepackage{framed} 
\definecolor{shadecolor}{rgb}{0.94, 0.97, 1.0}

\usepackage{url}

\usepackage[resetlabels]{multibib}
\usepackage{multirow}
\usepackage{multicol}
\usepackage{enumitem}
\usepackage[normalem]{ulem}

\usepackage{threeparttable}
\usepackage{adjustbox}
\usepackage{wrapfig}

\definecolor{cite_color}{HTML}{114083}
\definecolor{link_color}{RGB}{153, 0,0}  
\definecolor{url_color}{RGB}{153, 102,  0}
\definecolor{emp_color}{RGB}{0,0,255}
\definecolor{aliceblue}{rgb}{0.94, 0.97, 1.0}
\hypersetup{
 colorlinks,
 citecolor=cite_color,
 linkcolor=link_color,
 urlcolor=url_color}
\usepackage{caption}
\usepackage{pifont}
\usepackage{makecell}
\usepackage{subfigure} 
\usepackage{amsmath}
\usepackage{multirow}
\usepackage{multicol}
\usepackage{enumitem}
\usepackage{ulem}
\usepackage{threeparttable}
\usepackage{wrapfig}
\usepackage{scrextend}
\usepackage{thm-restate}
\usepackage{array,hhline}
\usepackage{amsmath,amsfonts}
\usepackage{tabularborder}
\usepackage[ruled]{algorithm2e}
\usepackage{setspace}
\usepackage{mathtools}
\DeclarePairedDelimiterX{\infdivx}[2]{(}{)}{%
  #1\;\delimsize\|\;#2%
}
\usepackage{cancel}
\newcommand{\infdiv}{D_\textrm{KL}\infdivx}
\newcommand\scalemath[2]{\scalebox{#1}{\mbox{\ensuremath{\displaystyle #2}}}}

\definecolor{mypink}{RGB}{255,211,150}
\newcommand{\up}{{\color{Tan}$\blacktriangle$}}

\newcommand{\down}{{\color{Gray}$\blacktriangledown$} }

\newcommand{\upbox}[1]{\colorbox{mypink}{#1}}
\newcommand{\revise}[1]{{#1}}
\newcommand{\reviseone}[1]{{#1}}

\SetKwComment{Comment}{$\triangleright$\ }{}
\usepackage{amssymb}
\usepackage{xspace}
\usepackage{wrapfig,tikz}
\usepackage{siunitx}

\usepackage{cleveref}
\newcommand{\CVGAE}{\text{CVGAE}\xspace}
\newcommand{\GraphDG}{\text{GraphDG}\xspace}
\newcommand{\CGCF}{\text{CGCF}\xspace}
\newcommand{\ConfVAE}{\text{ConfVAE}\xspace}
\newcommand{\ConfGF}{\text{ConfGF}\xspace}

\newcommand{\GeoMol}{\text{GeoMol}\xspace}

\newcommand{\GeoDiff}{\text{GeoDiff}\xspace}
\newcommand{\method}{\text{SubGDiff}\xspace}

\newcommand{\twoD}{\text{2D}}
\newcommand{\threeD}{\text{3D}}

\def\X{{\mathbf{X}}}
\usepackage{lmodern}
\usepackage{anyfontsize}
\usepackage[ruled]{algorithm2e}

\usepackage{amsmath,amsfonts,bm}

\newcommand{\mbf}[1]{\mathbf{#1}}
\newcommand{\mbb}[1]{\mathbb{#1}}

\newcommand{\mcal}[1]{\mathcal{#1}}

\def\1{\bm{1}}

\def\s{{\mathbf{s}}}

\def\R{{\mathbf{R}}}
\def\rmS{{\mathbf{S}}}

\def\vx{{\bm{x}}}
\def\vy{{\bm{y}}}

\def\mH{{\bm{H}}}

\DeclareMathAlphabet{\mathsfit}{\encodingdefault}{\sfdefault}{m}{sl}
\SetMathAlphabet{\mathsfit}{bold}{\encodingdefault}{\sfdefault}{bx}{n}

\def\gC{{\mathcal{C}}}

\def\gG{{\mathcal{G}}}

\def\gN{{\mathcal{N}}}

\newcommand{\E}{\mathbb{E}}

\DeclareMathOperator*{\argmin}{arg\,min}

\newcommand{\lowbound}[1]{\lfloor{#1}\rfloor}

\icmltitlerunning{\method: A Subgraph Diffusion Model to Improve Molecular Representation Learning}

\begin{document}

\twocolumn[
\icmltitle{\method: A Subgraph Diffusion Model to Improve Molecular Representation Learning}





\begin{icmlauthorlist}
\icmlauthor{Jiying Zhang}{comp}
\icmlauthor{Zijing Liu}{comp}
\icmlauthor{Yu Wang}{comp}
\icmlauthor{Yu Li}{comp}
\end{icmlauthorlist}

\icmlaffiliation{comp}{International Digital Economy Academy (IDEA)}

\icmlcorrespondingauthor{Z. Liu, Y. Li}{$\{$liuzijing, liyu$\}$@idea.edu.cn}


\vskip 0.3in
]

\printAffiliationsAndNotice{}  

\begin{abstract}
Molecular representation learning has shown great success in advancing AI-based drug discovery. 
The core of many recent works is based on the fact that the 3D geometric structure of molecules provides essential information about their physical and chemical characteristics. 
Recently, denoising diffusion probabilistic models have achieved impressive performance in 3D molecular representation learning.
However, most existing molecular diffusion models treat each atom as an independent entity, overlooking the dependency among atoms within the molecular substructures.
This paper introduces a novel approach that enhances molecular representation learning by incorporating substructural information within the diffusion process.
We propose a novel diffusion model termed \method for involving the molecular subgraph information in diffusion. Specifically, \method adopts three vital techniques: i) subgraph prediction, ii) expectation state, and iii) $k$-step same subgraph diffusion, to enhance the perception of molecular substructure in the denoising network.
Experimentally, extensive downstream tasks 
demonstrate the superior performance of our approach. The code is available at \href{https://github.com/youjibiying/SubGDiff}{Github}.
\end{abstract}

\section{Introduction}

Molecular representation learning (MRL) has attracted tremendous attention due to its significant role in learning from limited labeled data for applications like AI-based drug discovery~\citep{shen2019molecular,ijcai2022p518,zhang2022hypergraph} and material science~\citep{pollice2021data}.
From the perspective of physical chemistry, the 3D molecular conformation is crucial to determine the properties of molecules and the activities of drugs~\citep{cruz2014conformational}.
This has spurred the development of numerous geometric neural network architectures and self-supervised learning strategies aimed at leveraging 3D molecular structures to enhance performance on downstream molecular property prediction tasks~\citep{schutt2017schnet,zaidi2022pre_deepMind,liu2023group_molecularsde}.

Diffusion probabilistic models (DPMs) have shown remarkable power to generate realistic samples, especially in synthesizing high-quality images and videos~\citep{sohl2015deep,ho2020denoising}. By modeling the generation as a reverse diffusion process, DPMs transform a random noise into a sample in the target distribution. Recently, diffusion models have been successfully applied to molecular 3D conformation generation~\citep{xu2022geodiff, jing2022torsional}. 
The training process in DPMs, which involves reconstructing the original conformation from a noisy version across varying time steps, naturally lends itself to self-supervised representation learning~\citep{pan2023masked}.
Inspired by this, several works have used this technique for molecule pretraining~\citep{liu2023molecular_geossl,zaidi2022pre_deepMind}. 
Despite considerable progress, the full potential of DPMs in molecular representation learning remains underexplored. This lead us to investigate the question: \textit{Can we effectively enhance MRL
with the denoising network (noise predictor) of DPM? If yes, how to achieve it?}

\begin{figure}
    \centering

\includegraphics[width=0.85\linewidth]{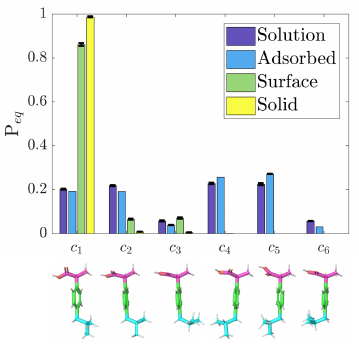}
\caption{\footnotesize Equilibrium probability of the six conformers (c1–c6) of the same molecule ibuprofen~(C13H18O2) in four different conditions. The 3D substructure is a significant characteristic of a molecule. {\scriptsize (Adapted with permission from ~\cite{marinova2018dynamics}. Copyright {2018} American Chemical Society.)}
}
\label{fig:intro}
\vspace{-2mm}
\end{figure}

To address this question, we first identify the gap between the current DPMs and the characteristics of molecular structures. 
Most diffusion models on molecules propose to independently inject continuous Gaussian noise into the every node feature~\citep{hoogeboom2022equivariant} or atomic coordinates of 3D molecular geometry~\citep{xu2022geodiff, zaidi2022pre_deepMind}. 
However, this approach treats each atom as an individual particle, overlooking the substructure within molecules, which is pivotal in molecular representation learning~\citep{yu2022molecular,wang2022improving,miao2022interpretable}.
As shown in~\autoref{fig:intro},
the 3D geometric substructure contains crucial information about the properties, such as the equilibrium distribution, crystallization and solubility~\citep{marinova2018dynamics}.
As a result, uniformly adding same-scale Gaussian noise to all atoms makes it difficult for the denoising network to capture the properties related to the substructure. So here we try to answer the previous question by designing a DPM involving the knowledge of substructures.



\begin{figure*}[htp]
    \centering
    \vspace{-0mm}
    \includegraphics[width=0.9\linewidth]{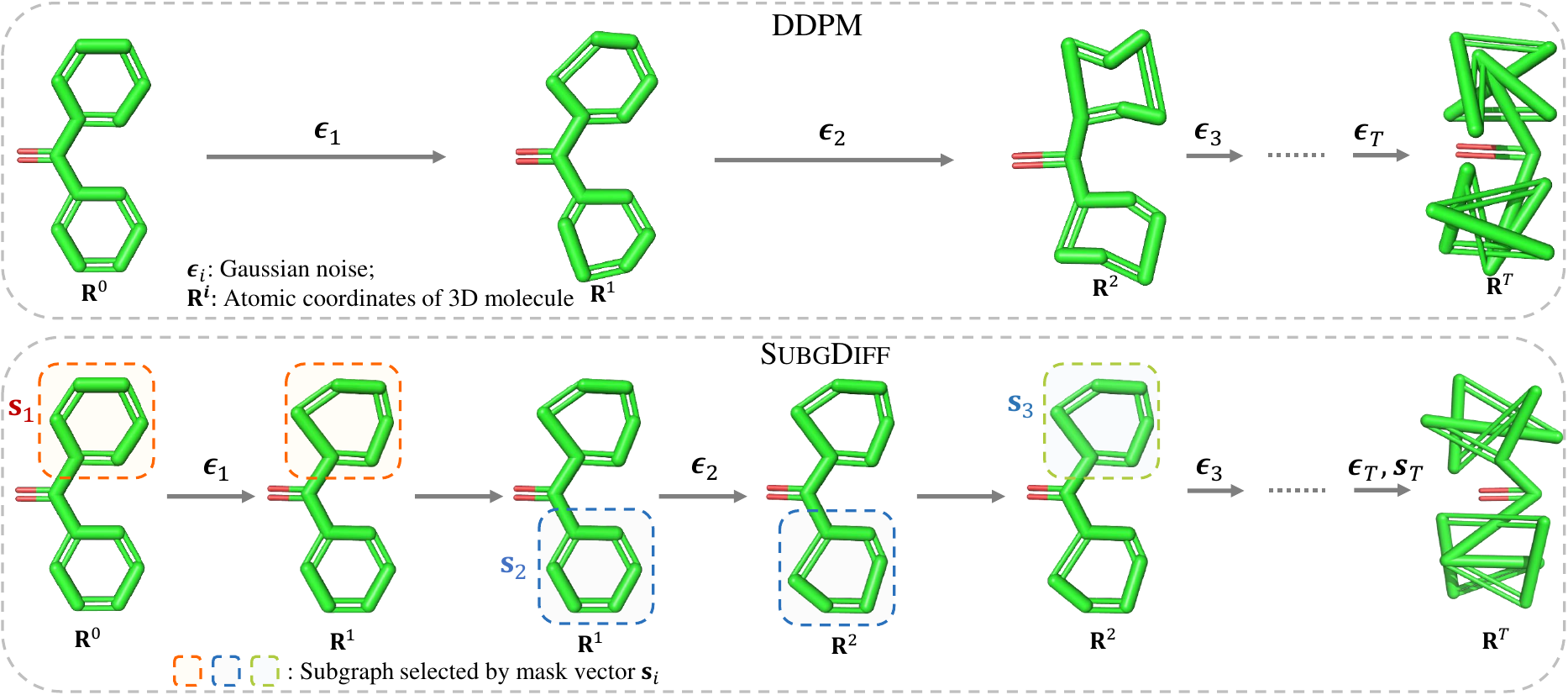}
    \caption{\footnotesize Comparison of forward process between DDPM~\citep{ho2020denoising} and subgraph diffusion.  For each step, DDPM adds noise into all atomic coordinates, while subgraph diffusion selects a subset of the atoms to diffuse.}
    \label{fig:ddpm_mask_diff}
    \vspace{-3mm}
\end{figure*}

Toward this goal, we propose a novel diffusion model termed \method, adding distinct Gaussian noise to different substructures of 3D molecular conformation. 
Specifically, instead of adding the same Gaussian noise to every atomic coordinate, \method introduces a discrete binary distribution to the diffusion process, where a mask vector sampling from the distribution can be used to select a subset of the atoms (i.e. subgraph) to determine which substructure the noise should be added to at the current time step (\autoref{fig:ddpm_mask_diff}). More importantly, 
in the training phase of \method, a subgraph prediction loss, resembling a node classifier, is integrated. This intentional inclusion explicitly directs the denoising network to capture substructure information from the molecules. Additionally, \method employs the \textit{expectation state diffusion} process to bolster its sampling capability and incorporates the \textit{k-step same-subgraph diffusion} process to optimize the model.

With the ability to capture the substructure information from the noisy 3D molecule, the denoising networks tend to gain more representation power.
The experiments on various 2D and 3D molecular property prediction tasks demonstrate the superior performance of our approach. To summarize, our contributions are as follows:
    (1) we incorporate the substructure information into diffusion models to improve molecular representation learning; 
    (2) we propose a new diffusion model \method that adopts subgraph prediction, expectation state and $k$-step same-subgraph diffusion to improve its sampling and training;  
    (3) the proposed representation learning method achieves superior performance on various downstream tasks.

\section{Related work}\label{sec:related}
\textbf{Diffusion models on graphs.} 
The diffusion models on graphs can be mainly divided into two categories: continuous diffusion and discrete diffusion. Continuous diffusion applies a Gaussian noise process on each node or edge ~\citep{ingraham2019generative,niu2020permutation}, including \GeoDiff~\citep{xu2022geodiff}, EDM~\citep{hoogeboom2022equivariant}, SubDiff~\cite{anonymous2024subdiff}. Meanwhile, discrete diffusion constructs the Markov chain on discrete space, including Digress~\citep{haefeli2022diffusion} and GraphARM~\citep{kong2023Autoregressive}.
However, it remains open to exploring fusing the discrete characteristic into the continuous Gaussian on graph learning, although a closely related work has been proposed for images and cannot be used for generation~\citep{pan2023masked}.
Our work, \method,
is the first diffusion model fusing subgraph, combining discrete characteristics and the continuous Gaussian.

\textbf{Conformation generation.}
Various deep generative models have been proposed for conformation generation, including 
\CVGAE~\citep{mansimov2019molecular},  \GraphDG~\citep{simm2020GraphDG}, \CGCF~\citep{xu2021cgcf},
\ConfVAE~\citep{xu2021end},  
\ConfGF~\citep{shi2021learning} and \GeoMol \citep{ganea2021geomol}.
Recently, diffusion-based methods have shown competitive performance.
Torsional Diffusion~\citep{jing2022torsional} raises a diffusion process on the hypertorus defined by torsion angles.
However, it is not suitable as a representation learning technique due to the lack of local information (length and angle of bonds).
\GeoDiff \citep{xu2022geodiff} generates molecular conformation with a diffusion model on atomic coordinates.
However, it views the atoms as separate particles, without considering the dependence between atoms from the substructure.

\textbf{SSL for molecular property prediction.}
There exist several works leveraging the 3D molecular conformation to boost the representation learning, including GraphMVP~\cite{liu2022pretraining},~GeoSSL~\citep{liu2023molecular_geossl}, the denoising pretraining approach raised by ~\citet{zaidi2022pre_deepMind} and MoleculeSDE~\citep{liu2023group_molecularsde}, etc. 
However, those studies have not considered the molecular substructure in the pertaining. 
In this paper, we concentrate on how to boost the perception of molecular substructure in the denoising networks through the diffusion model.

The discussion with more related works~(e.g. MDM~\citep{pan2023masked}, MDSM~\citep{lei2023masked} and SSSD~\citep{alcaraz2022diffusion}) can be found in Appendix \ref{sec:app_related_work}.

\section{Preliminaries}
\textbf{Notations.}
We use $\mbf I$ to denote the identity matrix with dimensionality implied by context, $\odot$ to represent the element product, and $\text{diag}(\s)$ to denote the diagonal matrix with diagonal elements of the vector $\s$. If not specified, both $\epsilon$ and $z$ represent noise sampled from the standard Gaussian distribution $\mcal N(\mbf 0,\mbf I)$.
The topological molecular graph can be denoted as $\mcal G ( \mcal V, \mcal E,\mbf X)$ where $\mcal V$ is the set of nodes, $\mcal E$ is the set of edges, $\mbf X$ is the node feature matrix, and its corresponding 3D Conformational Molecular Graph 
is represented as $G_{3D}(\mcal G,\mbf R)$, where $\mbf R = [{R}_1,\cdots, {R}_{|\mcal V|} ]\in\mbb R^{|\mcal V|\times 3}$ is the set of 3D coordinates of atoms.

\textbf{DDPM.} Denoising diffusion probabilistic models (DDPM)~\citep{ho2020denoising} is a typical diffusion model~\citep {sohl2015deep} which consists of a diffusion (aka forward) and a reverse process. 
In the setting of molecular conformation generation, the diffusion model adds noise on the 3D molecular coordinates $\mbf R$~\citep{xu2022geodiff}.

\textbf{Forward and reverse process.} Given the fixed variance schedule $\beta_1, \beta_2,\cdots,\beta_T$, the posterior distribution $q(R^{1:T}|R^{0})$ that is fixed to a Markov chain can be written as
\vspace{-2mm}
\begin{align}
    &\scalemath{0.88}{q(\R^{1:T}|\R^{0}) = \prod_{t=1}^{T} q(\R^{t}|\R^{t-1}),}
    \\
    &\scalemath{0.88}{q(\R^{t}|\R^{t-1})=\mcal{N}(\R^{t}, \sqrt{1-\beta_t } \R^{t-1}, \beta_t \mbf I)}.
\end{align}
To simplify notation, we consider the diffusion on single atom coordinate $R_v $ and omit the subscript $v$ to get the general notion $R$ throughout the paper. Let $\alpha_t = 1-\beta_t$, $\bar\alpha_t = \prod_{i=1}^{t}(1-\beta_t)$, and then the sampling of $R^t$ at any time step $t$ has the closed form: $  q(R^{t}|R^{0})=\mcal{N}(R^{t}, \sqrt{\bar\alpha_t } R^{0},(1-\bar\alpha_t)\mbf I)$.

The reverse process of DDPM is defined as a Markov chain starting from a Gaussian distribution $p(R^T) = \mcal N(R^T; \mbf 0, \mbf I)$:
\vspace{-2mm}
\begin{align}
    \scalemath{0.88}{p_\theta(R_{0:T})=p(R^T) \prod_{t=1}^{T} p_{\theta}(R^{t-1}|R^{t}) },\\
    \scalemath{0.88}{\quad p_{\theta}(R^{t-1}|R^{t}) = \mcal N (R^{t-1}; \mu_{\theta}(R^t,t), \sigma_{t}) },
\end{align}
where $\sigma_t=\frac{1-\bar\alpha_{t-1}}{1-\bar\alpha_t}\beta_t$ denote time-dependent constant. In DDPM, $\mu_\theta(R^t,t)$ is parameterized as $\mu_\theta(R^t,t) = \frac{1}{\bar\alpha_t}(R^t-\frac{\beta_t}{\sqrt{1-\bar\alpha_t}}\epsilon_\theta(R^t,t))$
and $\epsilon_{\theta}$, i.e., the \textit{denoising network}, is parameterized by a neural network where the inputs are $R^{t}$ and time step $t$. 

\textbf{Training and sampling.}
The training objective of DDPM is:
\vspace{-2mm}
\begin{align}
\label{eq:ddpm_simple_loss}
    \scalemath{0.88}{\mcal L_{simple}(\theta) = \mbb E_{t,R^0, 
    \epsilon}[\|\epsilon-\epsilon_{\theta }(\sqrt{\bar\alpha_t}R^0 + \sqrt{1-\bar\alpha_t}\epsilon, t)\|^2] }.
\end{align}

 After training, samples are generated through the reverse process $p_{\theta}(R^{0:T} )$.
Specifically, $R^T$ is first sampled from $\mcal N(\mbf 0, \mbf I)$, and $R^t$ in each step is predicted as follows,
\vspace{-2mm}
\begin{align}
\label{eq:ddpm_sampling}
    \scalemath{0.88}{R^{t-1} = \frac{1}{\sqrt{\alpha_t}}(R^t - \frac{1-\alpha_t}{\sqrt{1-\bar\alpha_t}}\epsilon_{\theta }(R^t,t))+ \sigma_t z, \quad  z \sim \mcal N(\mbf 0, \mbf I)}.
\end{align}

\section{\method}
Directly using DDPM on atomic coordinates of 3D molecules means each atom is viewed as an independent single data point. However, the substructures play an important role in molecular generation~\citep{jin2020hierarchical} and representation learning~\citep{zang2023hierarchical}. Ignoring the inherent interactions between atoms may hurt the ability of the denoising network to capture molecular substructure. In this paper, we propose to involve a mask operation in each diffusion step, leading to a new diffusion \method for molecular representation learning. Each mask corresponds to a subgraph in the molecular graph, aligning with the substructure in the 3D molecule. Furthermore, we incorporate a subgraph predictor and reset the state of the Markov Chain to the expectation of atomic coordinates, thereby enhancing the effectiveness of \method in sampling.
Additionally, we also propose $k$-step same-subgraph diffusion for training to effectively capture the substructure information.

\begin{wrapfigure}{R}{0.450\linewidth}
    \centering
            \vspace{-4mm}
    \includegraphics[width=0.95\linewidth]{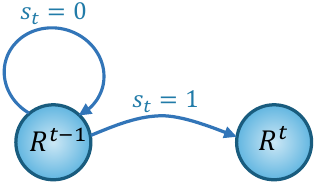}
    \caption{ The Markov Chain of \method is a lazy Markov Chain.}    \label{fig:markov_chain}
    \vskip - 0.1 in
\end{wrapfigure}

\vspace{-3mm}

\subsection{Involving subgraph into diffusion process} \label{subsec:maskedDiff}
In the forward process of DDPM, we have $R_v^t =  \sqrt{1- \beta_t } R_v^{t-1} + \sqrt{\beta_t} \epsilon_{t-1}, \forall v\in \mcal V$, in which the Gaussian noise $\epsilon_{t-1}$ is injected to every atom. Moreover, the training objective in \autoref{eq:ddpm_simple_loss} shows that the denoising networks would always predict a Gaussian noise for all atoms. Neither the forward nor reverse process of DDPM takes into account the substructure of the molecule. Instead, in \method, a mask vector $\mbf s_t=[s_{t_1},\cdots,s_{t_{|\mcal V|}}]^\top \in \{0,1\}^{|\mcal V|}$ is introduced to determine which atoms will be added noise at step $t$. The mask vector $\mbf s_t$ is sampled from a discrete distribution $p_{\s_t}(\mcal S \mid \mcal G)$ to select a subset of the atoms.
In molecular graphs, the discrete mask distribution $p_{\s_t}(\mcal S \mid \mcal G)$ is equivalent to the subgraph distribution, defined over a predefined sample space $\chi =\{G^i_{\text{sub}}\}_{i=1}^{N}$, where each sample is a connected subgraph extracted from $G$. Further, the distribution $p_{\s_t}(\mcal S \mid \mcal G)$ should keep the selected connected subgraph to cohere with the molecular substructures. Here, we adopt a Torsional-based decomposition method~\citep{jing2022torsional} (Details in Appendix~\ref{subsec:mask_distribution}). With the mask vector as latent variables $\s_{1:t}$, the state transition of the forward process can be formulated as (\autoref{fig:markov_chain}):  
\begin{align}
 R^t_v =
 \begin{cases}
   \sqrt{1- \beta_t } R_v^{t-1}   + \sqrt{\beta_t} \epsilon_{t-1}   &  \text{ if } s_{t_v}=1 \\
  R^t_v& \text{ if } s_{t_v} =0,
\end{cases}
\end{align}
which can be rewritten as $ R^t_v = \sqrt{1- s_{t_v}\beta_t } R_v^{t-1}   + \sqrt{s_{t_v}\beta_t}\epsilon_{t-1}  $.
The posterior distribution $q(R^{1:T}|R^{0},s_{1:T})$ can be expressed as matrix form:
\vspace{-2mm}
\begin{align}
    &\scalemath{0.88}{q(\R^{1:T}|\R^{0},\s_{1:T}) = \prod_{t=1}^{T} q(\R^{t}|\R^{t-1}, \s_t)}, \\
    & \scalemath{0.88}{q(\R^{t}|\R^{t-1},\s_t)=\mcal{N}(\R^{t}, \sqrt{1-\beta_t \text{diag}(\s_t) } \R^{t-1},\beta_t \text{diag}(\s_t) \mbf I)}\label{eq:1-step_transfer}.
\end{align}
To simplify the notation, we consider the diffusion on a single node $v$ and omit the subscript $v$ in $R^t_v$ and $s_{t_v}$ to get the notion $R^t$ and $s_t$. By defining $\gamma_t: = 1-s_t\beta_t$, $\bar\gamma_t: = \prod_{i=1}^{t}(1-s_t\beta_t)$, the closed form of sampling $R^t$ given $R^0$ is
\vspace{-2mm}
\begin{align}
\label{eq:R0ToRt}
    q(R^{t}|R^{0},s_{1:t})=\mcal{N}(R^{t}, \scalemath{0.88}{\sqrt{\bar\gamma_t } R^{0},(1-\bar\gamma_t)\mbf I)}.
\end{align}

\subsection{Reverse process learning}
The reverse process is decomposed as follows:
\vspace{-2mm}
\begin{align}
\label{eq:subgdiff_reverse}
\scalemath{0.88}{
    p_{\theta,\vartheta}(R^{0:T},s_{1:T}) = p(R^T)\prod_{t=1}^{T}p_{{\theta}}(R^{t-1}|R^t,s_t) p_\vartheta(s_{t}|R^t)},
\end{align}
where $p_{{\theta}}(R^{t-1}|R^t,s_t)$ and $ p_\theta(s_{t}|R^t)$ are both learnable models. In the context of molecular learning, the model can be regarded as first predicting which subgraph $\s_t$ should be denoised and then using the noise prediction network $p_{{\theta}}(R^{t-1}|R^t,s_t)$ to denoise the node position in the subgraph.

However, it is tricky to generate a 3D structure by adopting the typical training and sampling method used in \citet{ho2020denoising}. Specifically, following \citet{ho2020denoising}, the reverse process can be optimized by maximizing the variational lower bound (VLB) of $\log p(R^0)$ as follows,
\vspace{-2mm}
\begin{equation}
\label{eq:elob_simple_subgdiff}
\begin{aligned}
&\scalemath{0.88}{\log p(R^0)}
\geq  \scalemath{0.88}{\sum_{t = 1}^{T}\underbrace{\mathbb{E}_{q(R^{t}, s_{t}|R^0)}\left[\log \frac{p_{\vartheta}(s_{t}|R^t)}{ q(s_t)}\right]}_\text{subgraph prediction term}} + \\
&\quad\quad \quad\quad \scalemath{0.88}{ \underbrace{\mathbb{E}_{q(R^1,s_1|R^0)}\left[\log p_{{\theta}}(R^0|R^1)\right]}_\text{reconstruction term} -}
\\& \quad\quad \quad\quad\scalemath{0.88}{ \underbrace{\mathbb{E}_{q(s_{1:t})}\infdiv{q(R^T|R^0,s_{1:T})}{p(R^T)}}_\text{prior matching term}} -\\
& \scalemath{0.88}{\sum_{t=2}^{T} \underbrace{\mathbb{E}_{q(R^{t},s_{1:t}|R^0)}\left[\infdiv{q(R^{t-1}|R^t, R^0, s_{1:t})}{p_{{\theta}}(R^{t-1}|R^t,s_t)}\right]}_\text{denoising matching term}}. 
\end{aligned}
\end{equation}
\begin{figure*}
    \centering
    \vspace{-0.2cm}
    \includegraphics[width=0.98\linewidth]{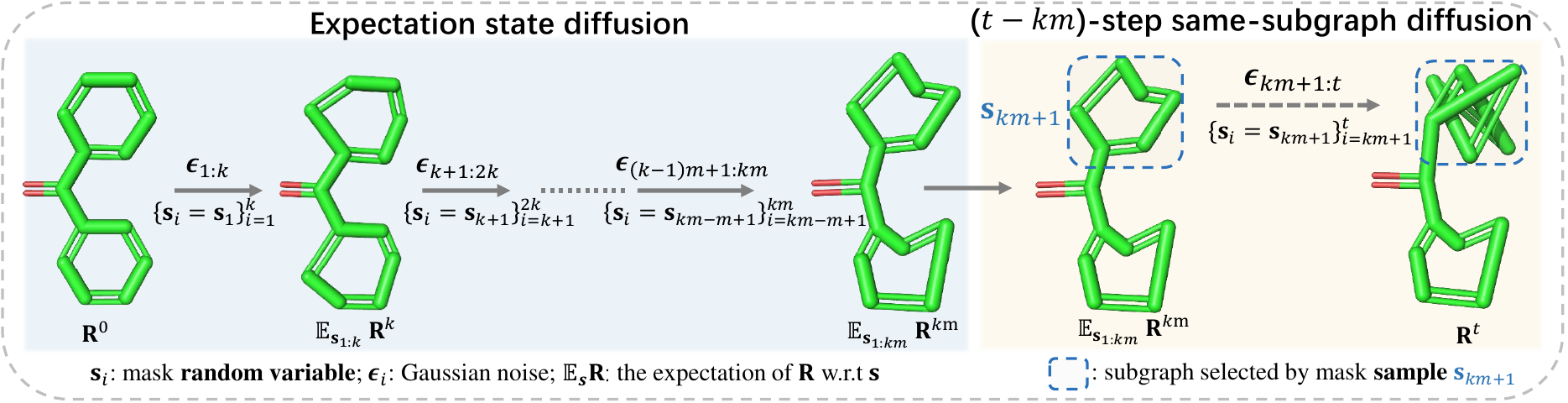}
    \vspace{-2mm}
     \caption{\revise{The forward process of \method.
     The state $0$ to $km$ uses the expectation state and the mask variables are the same in the interval $[ki,ki+k],i=0,1,...,m-1$.
     The state $km+1$ to  $t$ applies the same subgraph diffusion.} }
    \label{fig:subgraph_diff}
        \vspace{-3mm}
\end{figure*}
Details of the derivation are provided in the Appendix~\ref{sec:app_c1}.
The subgraph predictor $p_\vartheta(s_t| R^t)$ in the first term can be parameterized by a node classifier $s_{\vartheta}$.
For the denoising matching term that is closely related to sampling, by Bayes rule, the posterior $q(R^{t-1} | R^{t}, R^0, s_{1:t})$ can be written as:
\vspace{-2mm}
\begin{align}
\label{eq:bayes_simple_subGdiff} 
&\scalemath{0.88}{q(R^{t-1} | R^{t}, R^0, s_{1:t}) \propto \mathcal{N}(R^{t-1} ; \mu_q(R^t, R^0, s_{1:t}), \sigma^2_q(t) ) }, \\
&\scalemath{0.88}{\mu_q(R^t, R^0, s_{1:t}) = \frac{1}{\sqrt{1-\beta_ts_t}}(R^t - \scalemath{0.88}{\frac{\beta_t s_t}{\sqrt{(1-\beta_ts_t)(1-\bar\gamma_{t-1})+\beta_ts_t}}})},\notag
\end{align}
where $\sigma_q(t) $ is the standard deviation and $s_{1:t-1}$ are contained in $\bar\gamma_{t-1}$. 
Following DDPM and parameterizing $p_\theta(R^{t-1}|R^t,s_t)$  as $\mcal N(R^{t-1}; \mu_q(R^t, R^0, s_{1:t}) \epsilon_\theta(R^t,t) ), \sigma_q(t)\textbf{I})$, the training objective is
\vspace{-2mm}
\begin{align}
   \scalemath{0.88}{\mcal L_{simple}(\theta, \vartheta ) } &=
   \scalemath{0.88}{\mbb E_{t,\R^0,\s_t, 
    \epsilon}[ \|\text{diag}({\s_t})(\epsilon-\epsilon_{\theta }(\mcal G, \R^t, t))\|^2 } \notag \\
    &+  \scalemath{0.88}{
    \lambda \mathrm{BCE}(\s_t, \mbf s_\vartheta(\mcal G, \R^t,t))]}\label{eq:mask_predict_loss},
\end{align}
where 
$\mathrm{BCE}(\s_{t}, \mbf s_{\vartheta})$
is the binary cross entropy loss, $\lambda$ is the weight used for the trade-off, and $\mbf s_\vartheta $ is the subgraph predictor implemented as a node classifier with $G_{3D}(\mcal G,\mbf R^t)$ as input and shares a molecule encoder with $\epsilon_\theta$. The BCE loss employed here uses the subgraph selected at time-step $t$ as the target, thereby explicitly compelling the denoising network to capture substructure information from molecules.
Eventually, the $\mbf s_\vartheta$ can be used to infer the mask vector $\hat\s_t=s_\vartheta(\mcal G, \hat\R^t,t)$ during sampling.
Thus, the sampling process is:
\vspace{-2mm}
\begin{align}
    \scalemath{0.88}{R^{t-1} = \mu_q(R^t, R^0, s_{1:t-1}, p_\theta(s_t| R^t)) \epsilon_{\theta}(R^t,t) + \sigma_q(t) z }.\label{eq:mask_sampling_1}
\end{align}
However, using \autoref{eq:mask_predict_loss} and  \autoref{eq:mask_sampling_1} directly for training and sampling faces two issues.
First, the inability to access $s_{1:t-1}$ in $\mu_q$ during the sampling process hinders the step-wise denoising procedure, posing a challenge to the utilization of conventional sampling methods in this context.
Inferring $s_{1:t-1}$ solely from $R^t$ using another model $p_{{\theta}}(s_{1:t-1}|R^t,s_t)$ is also difficult due to the intricate modulation of noise introduced in $R^t$ through multi-step Gaussian noising. 
Second, training the subgraph predictor with \autoref{eq:mask_predict_loss} is challenging.
To be specific, the subgraph predictor should be capable of perceiving the sensible noise change between time steps $t-1$ and $t$. However, the noise scale $\beta_t$ is relatively small when $t$ is small, especially if the diffusion step is large (e.g. 1000). As a result, it is difficult to precisely predict the subgraph.

Next, we design two techniques: \textit{expectation state diffusion} and \textit{$k$-step same subgraph diffusion}, to effectively tackle the above issues.

\subsection{Expectation state diffusion.}
To tackle the dilemma above, we first calculate the denoising term in a new way. To eliminate the effect of mask series $s_{1:t}$ and improve the training of subgraph prediction loss, we use a new lower bound of the denoising term as follows:
\vspace{-2mm}
\begin{align}
       & \scalemath{0.88}{ \mathbb{E}_{q(R^{t},s_{1:t}|R^0)}\left[\infdiv{q(R^{t-1}|R^t, R^0, s_{1:t})}{p_{{\theta}}(R^{t-1}|R^t,s_t)}\right]} \notag\\
         &\leq\scalemath{0.88}{ \mathbb{E}_{q(R^{t},s_{1:t}|R^0)}\left[\infdiv{\hat{q}(R^{t-1}|R^t, R^0, s_{1:t})}{p_{\bm{\theta}}(R^{t-1}|R^t,s_t)}\right]}, \notag
\end{align}
where $\hat{q}(R^{t-1} |  R^{t}, R^0, s_{1:t})$ is defined as
\vspace{-2mm}
\begin{align}
 & \scalemath{0.88}{
   \frac{q( R^{t}|\mbb E_{s_{1:t-1}} R^{t-1},  R^0, s_{t})q(\mbb E_{s_{1:t-1}} R^{t-1}|R^0)}{q(R^{t}|R^0, s_{1:t})}}. \notag
\end{align}
It is an approximated posterior that only relies on the \textit{expectation} of $R_t$ and $s_t$. 
This lower bound defines a new forward process, in which, state $0$ to state $t-1$ use the $\mbb E_{\s_{1:t-1}}\R^{t-1}$ and state $t$ remains as \autoref{eq:1-step_transfer}. Assume each node $v \in \mcal V$,  $s_{t_v} \sim Bern(p)$ (i.i.d. w.r.t. $t$). Formally, we have 
\begin{align}
&\scalemath{0.88}{q(R^t | R^{t-1}, s_t) = \mcal{N}(R^{t}; \sqrt{1-s_t\beta_t } \mbb E R^{t-1},(s_t\beta_t)\mbf I) }, \\
&\scalemath{0.85}{q(\mbb ER^{t-1} | R^0,s_{1:t-1}) =  \mcal{N}(\mbb E R^{t-1}; \prod_{i=1}^{t-1}\sqrt{\alpha_i} R^{0},  p^2\sum_{i=1}^{t-1} \prod_{j=i+1}^{t-1}\alpha_j \beta_{i} I)},
\label{eq:1_step_expection}
\end{align}
where $\alpha_i := (p\sqrt{1-\beta_i}+1-p)^2$ and $\bar{\alpha}_t := \prod_{i=1}^{t}\alpha_i$ are general form of $\alpha_j$ and $\bar\alpha_j$ in DDPM ($p=1$), respectively. Intuitively, this process is equivalent to using mean state $\mbb E R^{t-1}$ to replace $R^{t-1}$ during the forward process. This estimation is reasonable since the expectation $\mbb E_{s_{1:t-1}}R^{t-1} $ is like a cluster center of $R^{t-1}
 $, which can represent $R^{t-1}$ properly. 
Thus, the approximated posterior becomes
\begin{align}
    & \scalemath{0.88}{\hat{q}(R^{t-1} |  R^{t}, R^0, s_{1:t}) \propto \mathcal{N}(R^{t-1} ; \mu_{\hat q}(R^t, R^0, s_{1:t}), \sigma^2_{\hat{q}}(t) ) },
\end{align}
where 
\begin{align}
&\scalemath{0.8}{\mu_{\hat{q}}(R^t, R^0, s_{1:t}) := \frac{1}{\sqrt{1-\beta_ts_t}}(R^t - \scalemath{0.88}{\frac{s_t\beta_t}{\sqrt{(\hat{s}_t\beta_t + (1-s_t\beta_t)p^2\sum_{i=1}^{t-1} 
    \frac{\bar{\alpha}_{t-1}}{\bar{\alpha}_i} \beta_{i} }}})}\notag \\
   & \scalemath{0.88}{\sigma^2_{\hat{q}}(t) :={\hat{s}_t\beta_t p^2 \sum_{i=1}^{t-1} 
    \frac{\bar{\alpha}_{t-1}}{\bar{\alpha}_i} \beta_{i}}/
    {(\hat{s}_t\beta_t + p^2(1-\hat{s}_t\beta_t)\sum_{i=1}^{t-1} 
    \frac{\bar{\alpha}_{t-1}}{\bar{\alpha}_i} \beta_{i})}}. \notag
\end{align}

We parameterize $p_\theta(R^{t-1}|R^t,s_t)$  as $\mcal N(R^{t-1}; \mu_{\hat{q}}(R^t, R^0, s_{1:t}) \epsilon_\theta(R^t,t) ), \sigma_{\hat{q}}(t)\textbf{I})$, and adopt the same training objective as \autoref{eq:mask_predict_loss}. By employing the sampling method $R^{t-1} = \mu_{\hat{q}}(R^t, R^0, s_{1:t-1}, p_\theta(s_t| R^t)) \epsilon_{\theta}(R^t,t) + \sigma_{\hat{q}}(t) z $, we observe that the proposed expectation state enables a step-by-step execution of the sampling process.
\reviseone{Moreover, using expectation is beneficial to reduce the complexity of $R^t$ for predicting the mask $s_t$ during training. This will improve the denoising network to perceive the substructure when we use the diffusion model for self-supervised learning.}

\subsection{$k$-step same-subgraph diffusion.}
To reduce the complexity of the mask series $(\s_1,\s_2,\cdots,\s_T)$ and accumulate more noise on the same subgraph for facilitating the convergence of the subgraph prediction loss, we generalize the one-step subgraph diffusion to $k$-step same subgraph diffusion~(\autoref{fig:fix_mask_diff} in Appendix), in which the selected subgraph will be continuously diffused $k$ steps. After that, the difference between the selected and unselected parts will be distinct enough to help the subgraph predictor perceive it.
The forward process of $k$-step same subgraph diffusion can be written as ($t>k,k\in \mbb N$):
\begin{align}
    \label{eq:t-step_transfer_mask}
    \scalemath{0.82}{q(R^{t}|R^{t-k})=\mcal{N}\left(R^{t}, \sqrt{\prod_{i=t-k+1}^{t}(1-s_{t-k+1}\beta_i) } R^{t-k},\sigma^k_t \right)},
\end{align}
where ${\sigma^k_t  = (1-\prod_{i=t-k}^{t}(1-s_{t-k+1}\beta_i)\mbf I}$.

\subsection{Training and sampling of \method}
By combining the expectation state and $k$-step same-subgraph diffusion, \method first divides the entire diffusion step $T$ into $T/k$ diffusion intervals. In each interval $[ki,k(i+1)]$, the mask vectors $\{\s_j\}_{j=ki+2}^{k(i+1)}$ are equal to $\s_{ki+1}$.
\method then adopts the expectation state at the split time step $\{ik| i=1,2,\cdots\}$ to eliminate the effect of $\{\s_{ik+1}|i=1,2,\ldots\}$, that is, gets the expectation of $\mbb E\R^{ik}$ at step $ ik$ w.r.t. $\s_{ik+1}$.
 Overall, the diffusion process of \method is a two-phase diffusion process. In the first phase, the state $1$ to state $k\lowbound{t/k}$ use the expectation state diffusion, while in the second phase, state $k(\lowbound{t/k})+1$ to state $t$ use the $k$-step same subgraph diffusion. The state transition refers to~\autoref{fig:subgraph_diff}. With $m :=\lowbound{t/k}$, the two phases can be formulated as follows,

    \textbf{Phase I}: Step $0\to km$: 
        $\mbb E_{s_{1:km}} R^{km}=\sqrt{\bar\alpha_{m}} R^0 +  p\sqrt{\sum_{l=1}^{m} \frac{\bar\alpha_{m}}{\bar\alpha_{l}} (1-\prod_{i=(l-1)k+1}^{kl}(1-\beta_i))} \epsilon_0,$
    where $\alpha_j= (p\sqrt{\prod_{i=(j-1)k+1}^{kj}(1-\beta_i)} + 1-p)^2$ is a  general forms of $\alpha_j$ in \autoref{eq:1_step_expection} (in which case $k=1$) and $\bar{\alpha}_t = \prod_{i=1}^{t}\alpha_i$. 
    In the rest of the paper, $\alpha_j$ denotes the general version without a special statement. Actually, $\mbb E_{s_{1:km}} R^{km}$  only calculate the expectation w.r.t. random variables $\{\s_{ik+1}|i=1,2,\cdots\}$.
    
    \textbf{Phase II}: Step $km+1 \to t$: The phase is a $(t-km)$-step same mask diffusion. $R^t = \sqrt{\prod_{i=km +1}^{t} (1-\beta_i s_{km+1 })} \mbb E_{s_{1:km}} R^{km} + \sqrt{1- \prod_{i=km +1}^{t} (1-\beta_i s_{km+1  })} \epsilon_{km} $.

Let $\gamma_i = 1-\beta_i s_{k m+1}$, $\bar\gamma_t = \prod_{i=1}^t \gamma_i$, and $\bar\beta_t = \prod_{i=1}^{t}(1-\beta_i)$. We can drive the single-step state transition: $    \scalemath{0.8}{q(R^{t}|R^{t-1})=\mcal{N}(R^{t}; \sqrt{\gamma_t } R^{t-1},(1-\gamma_t)\mbf I)}$ and
\vspace{-2mm}
\begin{align}
&\scalemath{0.8}{q(R^{t-1}|R^0)} =\scalemath{0.8}{\mcal{N}(R^{t-1}; \sqrt{\frac{\bar\gamma_{t-1}\bar\alpha_{m}}{\bar\gamma_{km}}}R^0, 
 \delta I)}, \label{eq:Rt_R0}  \\ 
 &\scalemath{0.8}{\delta }:= \scalemath{0.8}{\frac{\bar\gamma_{t-1}}{\bar\gamma_{km}} p^2 \sum_{l=1}^{m}
    \frac{\bar\alpha_{m}}{\bar\alpha_{l}} 
    (1-\frac{\bar\beta_{kl}}{\bar\beta_{(l-1)k}}) + 
1-\frac{\bar\gamma_{t-1}}{\bar\gamma_{km}}}.
\label{eq:delta} 
\end{align}
Then we reuse the training objective in \autoref{eq:mask_predict_loss} as the objective of \method: 
\begin{align}
   \scalemath{0.88}{\mcal L_{simple}(\theta, \vartheta ) } &=
   \scalemath{0.88}{\mbb E_{t,\R^0,\s_t, 
    \epsilon}[ \|\text{diag}({\s_t})(\epsilon-\epsilon_{\theta }(\mcal G, \R^t, t))\|^2 } \notag \\
    &+  \scalemath{0.88}{
    \lambda \mathrm{BCE}(\s_t, s_\vartheta(\mcal G, \R^t,t))]}\label{eq:subgdiff_loss},
\end{align}
where $\R^t$ is calculated by \autoref{eq:Rt_R0}. 
 
\paragraph{Sampling.}
Although the forward process uses the expectation state w.r.t. $\s$, we can only update the mask $\hat\s_{t}$ at $t=ik,i=1,2,\cdots$ because sampling only needs to get a subgraph from the distribution in the $k$-step interval.
Eventually, adopting the $\delta$ defined in \autoref{eq:delta}, the sampling process is shown below,
\vspace{-2mm}
\begin{align}
\label{eq:SubGDiff_sampling}
    \scalemath{0.8}{R^{t-1} }
    &=\scalemath{0.8}{\frac{1}{\sqrt{ \gamma_t}}(R^t -\frac{ 
\hat{s}_{k m+1}\beta_t}{\sqrt{\gamma_t \delta + \hat{s}_{k m+1}\beta_t}} \epsilon_{\theta}(
R^t,t) )}\notag \\
    &\scalemath{0.8}{+ \frac{\sqrt{ \hat{s}_{k m+1}\beta_t} \sqrt{\frac{\bar\gamma_{t-1}}{\bar\gamma_{km}} p^2 \sum_{l=1}^{m}
    \frac{\bar\alpha_{m}}{\bar\alpha_{l}} 
    (1-\frac{\bar\beta_{kl}}{\bar\beta_{(l-1)k}}) + 
1-\frac{\bar\gamma_{t-1}}{\bar\gamma_{km}}}}{\sqrt{\gamma_t \delta + \hat{s}_{k m+1}\beta_t}}   z},
\end{align}
where $ z\sim \mcal N(\mbf 0, \mbf I), \,m=\lowbound{t/k}$ and $\hat{s}_{km+1} = s_{\vartheta}(\mcal G,R^{km+k}, km+k)$.
The subgraph selected by $\hat{\s}_{km+1}$  will be generated in from the steps $km+k$ to $km$. The mask predictor can be viewed as a discriminator of important subgraphs, indicating the optimal subgraph should be recovered in the next $k$ steps. After one subgraph (substructure) is generated properly, the model can gently fine-tune the other parts of the molecule (c.f. the video in supplementary material). This subgraph diffusion would intuitively increase the robustness and generalization of the generation process, which is also verified by the experiments in sec. \ref{subsec:Confor_gene}. 
The training and sampling algorithms of \method are summarized in Alg.~\ref{alg:trainingSubGDiff} and Alg.~\ref{alg:SamplingSubGDiff}.


\begin{algorithm}[htp]
\setstretch{1.0}
\caption{Training \method } \label{alg:trainingSubGDiff}
\KwIn{A molecular graph $G_{3D}$, $k$ for $k$-step same-subgraph diffusion}
Sample $t \sim \mathcal U(1, ..., T)$ , $\epsilon \sim \mcal N(\mbf 0, \mbf I)$ \\
Sample $\mbf s_{k\lowbound{t/k}+1}  \sim  p_{s_{k\lowbound{t/k}+1}}(\mcal S \mid \mcal G)$ \\
$\R^t \gets  q(\R^t| \R^{0})$  \Comment*[f]{\autoref{eq:Rt_R0}} \\
$\mcal L \gets$ \autoref{eq:subgdiff_loss} \\
$\operatorname{optimizer.step}(\mcal L)$ \\
\end{algorithm}
\vspace{-2mm}

\vspace{-2mm}

\begin{algorithm}[htbp]
\setstretch{1.}
\caption{Sampling from \method }\label{alg:SamplingSubGDiff}
$k$ is the same as training, for $k$-step same-subgraph diffusion;
Sample $\R^T  \sim \mcal N(\mbf 0, \mbf I)$;  
\\
\For{t = T \KwTo 1}{
    $\mbf z  \sim \mcal N(\mbf 0, \mbf I)$ if $t>1$, else $\mbf z =\mbf 0$  \\
    \textbf{If} $t\%k==0$  or $t==T$: $\hat \s \gets s_\vartheta(\mcal G,\R^{t},t)$ \\
    $ \hat\epsilon \gets \epsilon_\theta (\mcal G,\R^{t},t)$ \Comment*[f]{Posterior}\\
    $\R^{t-1} \gets$ \autoref{eq:SubGDiff_sampling} \Comment*[f]{sampling} 
}
\Return $\R^0$
\end{algorithm}
\vspace{-2mm}

\begin{table*}[tb]
\setlength{\tabcolsep}{5pt}
\fontsize{9}{9}\selectfont
\centering
\caption{
\small
Results (mean absolute error) on MD17 \textbf{force} prediction. The best and second best results are marked in bold and underlined.
}
\vspace{-2mm}
\label{tab:MD17_result}
\begin{adjustbox}{max width=\textwidth}
\begin{tabular}{l c c c c c c c c}
\toprule
Pretraining & Aspirin $\downarrow$ & Benzene $\downarrow$ & Ethanol $\downarrow$ & Malonaldehyde $\downarrow$ & Naphthalene $\downarrow$ & Salicylic $\downarrow$ & Toluene $\downarrow$ & Uracil $\downarrow$ \\
\midrule
-- (random init) & 1.203 & 0.380 & 0.386 & 0.794 & 0.587 & 0.826 & 0.568 & 0.773\\

Type Prediction & 1.383 & 0.402 & 0.450 & 0.879 & 0.622 & 1.028 & 0.662 & 0.840\\

Distance Prediction & 1.427 & 0.396 & 0.434 & 0.818 & 0.793 & 0.952 & 0.509 & 1.567\\

Angle Prediction & 1.542 & 0.447 & 0.669 & 1.022 & 0.680 & 1.032 & 0.623 & 0.768\\

3D InfoGraph & 1.610 & 0.415 & 0.560 & 0.900 & 0.788 & 1.278 & 0.768 & 1.110\\

RR & 1.215 & 0.393 & 0.514 & 1.092 & 0.596 & 0.847 & 0.570 & 0.711\\

InfoNCE & 1.132 & 0.395 & 0.466 & 0.888 & 0.542 & 0.831 & 0.554 & 0.664\\

EBM-NCE & 1.251 & 0.373 & 0.457 & 0.829 & 0.512 & 0.990 & 0.560 & 0.742\\

3D InfoMax & 1.142 & 0.388 & 0.469 & 0.731 & 0.785 & 0.798 & 0.516 & 0.640\\

GraphMVP & 1.126 & 0.377 & 0.430 & 0.726 & 0.498 & 0.740 & 0.508 & 0.620\\

Denoising & 1.364 & 0.391 & 0.432 & 0.830 & 0.599 & 0.817 & 0.628 & 0.607\\

GeoSSL & \underline{{1.107}} & 0.360 & 0.357 & 0.737 & 0.568 & 0.902 & {0.484} & 0.502\\

MoleculeSDE (VE) & {1.112} & \underline{{0.304}} & {\underline{0.282}} & {0.520} & {0.455} & {0.725} & 0.515 & \underline{{0.447}}\\

MoleculeSDE (VP) & 1.244 & {0.315} & {0.338} & {\underline{0.488}} & \underline{{0.432}} & \underline{{0.712}} & \underline{{0.478}} & {0.468}\\
\midrule
Ours & \bf 0.880 & \textbf{0.252} & \textbf{0.258} & \textbf{0.459} & {\textbf{0.325}} & {\textbf{0.572}} & {\textbf{0.362}} & \textbf{0.420}\\
\bottomrule
\end{tabular}
\end{adjustbox}
\end{table*}

\section{Experiments}\label{sec:experiment}
We conduct experiments to address the following two questions:
1) Can substructures improve the representation ability of the denoising network when using diffusion as self-supervised learning?
2) How does the proposed subgraph diffusion affect the generative ability of the diffusion models?
For the first question, we employ \method as a denoising pretraining task and evaluate the performances of the denoising network on various downstream tasks.
For the second one, we compare \method with the vanilla diffusion model \GeoDiff~\cite{xu2022geodiff} on the task of molecular conformation generation.

\begin{table*}[ht!]
\caption{\small{
Results for MoleculeNet (with 2D topology only). 
We report the mean (and standard deviation) ROC-AUC of three random seeds with scaffold splitting for each task. The backbone is GIN.
The best and second best results are marked  {bold} and {underlined}, respectively.}
}
\vspace{-2mm}
\label{tab:main_results_moleculenet_2D}
\centering
\begin{adjustbox}{max width=\textwidth}
\small
\setlength{\tabcolsep}{4pt}
\centering
\begin{tabular}{l c c c c c c c c c}
\toprule
Pre-training & BBBP $\uparrow$ & Tox21 $\uparrow$ & ToxCast $\uparrow$ & Sider $\uparrow$ & ClinTox $\uparrow$ & MUV $\uparrow$ & HIV $\uparrow$ & Bace $\uparrow$ & Avg $\uparrow$ \\
\midrule
-- (random init) & 68.1$\pm$0.59 & 75.3$\pm$0.22 & 62.1$\pm$0.19 & 57.0$\pm$1.33 & 83.7$\pm$2.93 & 74.6$\pm$2.35 & 75.2$\pm$0.70 & 76.7$\pm$2.51 & 71.60 \\

AttrMask & 65.0$\pm$2.36 & 74.8$\pm$0.25 & 62.9$\pm$0.11 & {61.2$\pm$0.12} & \underline{{87.7$\pm$1.19}} & 73.4$\pm$2.02 & 76.8$\pm$0.53 & 79.7$\pm$0.33 & 72.68 \\

ContextPred & 65.7$\pm$0.62 & 74.2$\pm$0.06 & 62.5$\pm$0.31 & \underline{62.2$\pm$0.59} & 77.2$\pm$0.88 & 75.3$\pm$1.57 & 77.1$\pm$0.86 & 76.0$\pm$2.08 & 71.28 \\
InfoGraph & 67.5$\pm$0.11 & 73.2$\pm$0.43 & 63.7$\pm$0.50 & 59.9$\pm$0.30 & 76.5$\pm$1.07 & 74.1$\pm$0.74 & 75.1$\pm$0.99 & 77.8$\pm$0.88 & 70.96 \\
MolCLR & 66.6$\pm$1.89 & 73.0$\pm$0.16 & 62.9$\pm$0.38 & 57.5$\pm$1.77 & 86.1$\pm$0.95 & 72.5$\pm$2.38 & 76.2$\pm$1.51 & 71.5$\pm$3.17 & 70.79 \\

3D InfoMax & 68.3$\pm$1.12 & 76.1$\pm$0.18 & 64.8$\pm$0.25 & 60.6$\pm$0.78 & 79.9$\pm$3.49 & 74.4$\pm$2.45 & 75.9$\pm$0.59 & 79.7$\pm$1.54 & 72.47 \\

GraphMVP & 69.4$\pm$0.21 & 76.2$\pm$0.38 & 64.5$\pm$0.20 & 60.5$\pm$0.25 & 86.5$\pm$1.70 & 76.2$\pm$2.28 & 76.2$\pm$0.81 & {79.8$\pm$0.74} & 73.66 \\

MoleculeSDE(VE) & {{68.3$\pm$0.25}} & {76.9$\pm$0.23} & \underline{64.7$\pm$0.06} &  60.2$\pm$0.29 & 80.8$\pm$2.53  & \underline{76.8$\pm$1.71} & {77.0$\pm$1.68} &  79.9$\pm$1.76 & {73.15} \\

MoleculeSDE(VP)  & \underline{70.1$\pm$1.35} & \underline{{77.0$\pm$0.12}} & {64.0$\pm$0.07} & 60.8$\pm$1.04 & {82.6$\pm$3.64} & 76.6$\pm$3.25  & \underline{77.3$\pm$1.31} & \underline{81.4$\pm$0.66} &\underline{{73.73}} \\

\midrule
Ours & \textbf{70.2$\pm$2.23} & \textbf{ 77.2$\pm$0.39} & \textbf{65.0$\pm$0.48 } & \textbf{62.2$\pm$0.97  } & \textbf{88.2$\pm$1.57 } & {\textbf{77.3$\pm$1.17}} & \textbf{77.6$\pm$0.51 } &\textbf{ 82.1$\pm$0.96} & {\textbf{74.85}} \\

\bottomrule
\end{tabular}
\end{adjustbox}
\end{table*}

\subsection{\method improves molecule representation learning for molecular property prediction}
To verify the introduced substructure in the diffusion can enhance the denoising network for representation learning, we pretrain with our \method objective and finetune on different downstream tasks.

\textbf{Dataset and settings.} For pretraining, we follow \cite{liu2023group_molecularsde} and use PCQM4Mv2 dataset~\citep{hu2020ogb}. It's a sub-dataset of PubChemQC~\citep{nakata2017pubchemqc} with 3.4 million molecules with 3D geometric conformations. 
We use various molecular property prediction datasets as downstream tasks. 
For tasks with 3D conformations, we consider the dataset MD17 and follow the literature~\citep{schutt2017schnet, schutt2021equivariant, liu2023molecular_geossl} of using 1K for training and 1K for validation, while the test set (from 48K to 991K) is much larger.

For downstream tasks with only 2D molecule graphs, we use eight molecular property prediction tasks from MoleculeNet~\citep{wu2017moleculenet}.

\textbf{Pretraining framework.}
To explore the potential of the proposed method for representation learning, we consider MoleculeSDE~\citep{liu2023group_molecularsde}, a SOTA pretraining framework, to be the training backbone, where \method is used for the $2D \to3D$ model and the mask operation is extended to the node feature and graph adjacency for the $3D \to 2D$ model. The details can be found in Appendix~\ref{appsubsec:setting}.

\textbf{Baselines.}
For 3D tasks, we incorporate two self-supervised methods [Type Prediction, Angle Prediction], and three contrastive methods [InfoNCE~\citep{oord2018representation} and EBM-NCE~\citep{liu2022pretraining} and 3D InfoGraph~\citep{liu2023molecular_geossl}]. Two denoising baselines are also included [GeoSSL~\citep{liu2023molecular_geossl}, Denoising~\cite{zaidi2022pre_deepMind} and MoleculeSDE].
For 2D tasks, the baselines are AttrMask, ContexPred~\citep{hu2020strategies},  InfoGraph~\citep{sun2019infograph},  MolCLR~\citep{wang2022molecular}, 3D InfoMax~\citep{stark20223d}, GraphMVP~\citep{liu2022pretraining} and MoleculeSDE.
For details, see Appendix \ref{appsec:SSL}.

\textbf{Results.} As shown in Table \ref{tab:MD17_result} and Table \ref{tab:main_results_moleculenet_2D}, \method outperforms MoleculeSDE in most downstream tasks. The results demonstrate that the introduced mask vector helps the perception of molecular substructure in the denoising network during pretraining. Further, \method achieves SOTA performance compared to all the baselines. This also reveals that the proposed \method objective is promising for molecular representation learning due to the involvement of the prior knowledge of substructures during training.
More results on the QM9 dataset~\citep{schutt2017schnet} can be found in Appendix ~\ref{appsubsec:3DQM9}.

\subsection{\method benefits conformation generation}
\label{subsec:Confor_gene}
We have proposed a new diffusion model to enhance molecular representing learning, where the base diffusion model (\GeoDiff) is initially designed for conformation generation. To evaluate the effects of \method on the generative ability of diffusion models, we evaluate its generation performance and generalization ability.

\begin{table*}[!t]
     \caption{Results for conformation generation on \textbf{GEOM-QM9} dataset with different diffusion timesteps. DDPM~\citep{ho2020denoising} is the sampling method used in \GeoDiff. Our proposed sampling method (Algorithm \ref{alg:SamplingSubGDiff}) can be viewed as a DDPM variant. {\color{mypink}{$\blacksquare$}} / {\color{Gray}{$\blacksquare$}}  denotes  \method outperforms/underperforms  \GeoDiff.
    }
    \label{tab:qm9}
     \vspace{-2mm}
    \centering
    \resizebox{1.0\textwidth}{!}{
    \begin{tabular}{lcl|llll|llll}
    \toprule[1.0pt]
     & & & \multicolumn{2}{c}{\shortstack[c]{COV-R (\%) $\uparrow$}} & \multicolumn{2}{c|}{\shortstack[c]{MAT-R (\si{\angstrom}) $\downarrow$ }}  & \multicolumn{2}{c}{\shortstack[c]{COV-P (\%) $\uparrow$}}  & \multicolumn{2}{c}{\shortstack[c]{MAT-P (\si{\angstrom}) $\downarrow$ }} \\
     Models & Timesteps& Sampling method &  Mean & Median & Mean & Median & Mean & Median & Mean & Median \\
    \midrule[0.8pt]
     \GeoDiff&     5000& DDPM &    80.36 & 83.82 
 &  0.2820 & 0.2799   &  53.66 &  50.85  & 0.6673   &  0.4214 \\
      
       \method & 5000&  DDPM (ours)  & \upbox{90.91}   & \upbox{95.59} & \upbox{0.2460} & \upbox{0.2351} & \colorbox{Gray}{50.16} & \colorbox{Gray}{48.01} &  \upbox{0.6114} & \colorbox{Gray}{0.4791}\\
    \midrule[0.5pt]
        \GeoDiff&   500& DDPM &  80.20   &   83.59& 0.3617& 0.3412 & 45.49    &  45.45 & 1.1518  &0.5087  \\
        \method& 500&  DDPM (ours) &  \upbox{89.78}	& \upbox{94.17}	&\upbox{0.2417}	&\upbox{0.2449}&	\upbox{50.03}&	\upbox{48.31}	&\upbox{0.5571}	&\upbox{0.4921}\\
                 \midrule[0.3pt]
 \GeoDiff&  200& DDPM  &     69.90    &   72.04& 0.4222 & 0.4272 & 36.71 &     33.51 & 0.8532 & 0.5554 \\
\method &  200& DDPM (ours) &       \upbox{85.53}   &    \upbox{88.99} &  \upbox{0.2994} & \upbox{0.3033} & \upbox{47.76}  &    \upbox{45.89}  & \upbox{0.6971} & \upbox{0.5118} \\
    \bottomrule[1.0pt]
    \end{tabular}
    }
    \vspace{-2mm}
\end{table*}

\begin{table}[!t]
    \caption{Results on the \textbf{GEOM-QM9} dataset for domain generalization. Except for \GeoDiff and \method, the other methods are trained with in-domain data.}
    \label{tab:qm9_DG}
    \vspace{-2mm}
    \centering
    \scalebox{0.65}{
    \begin{tabular}{lc|cccc}
    \toprule[1.0pt]
     & & \multicolumn{2}{c}{\shortstack[c]{COV-R (\%) $\uparrow$}} & \multicolumn{2}{c}{\shortstack[c]{MAT-R (\si{\angstrom}) $\downarrow$ }}   \\
    Models &  Train data & Mean & Median & Mean & Median  \\
    \midrule[0.8pt]
    \CVGAE~\citep{mansimov2019molecular} & QM9 & 0.09 & 0.00 & 1.6713 & 1.6088  \\ 
    \citet{simm2020GraphDG} & QM9 &  73.33 & 84.21 & 0.4245 & 0.3973 \\ 
    \CGCF~\citep{xu2021cgcf} & QM9 & 78.05 & 82.48 & 0.4219 & 0.3900  \\
    \ConfVAE~\citep{xu2021end} & QM9 & 77.84 & 88.20 & 0.4154 & 0.3739  \\
    \GeoMol~\citep{ganea2021geomol}  & QM9 & 71.26 & 72.00 & 0.3731 & 0.3731 \\ 
     \GeoDiff&  Drugs& 74.94 & 79.15 &  0.3492 &  0.3392   \\

      \midrule[0.3pt]
    \bf \method &  Drugs &\textbf{83.50} & \textbf{88.70} & \bf{0.3116} & \bf{0.3075} \\
    
    \bottomrule[1.0pt]
    \end{tabular}   }
    \vspace{-5mm}
\end{table}

\textbf{Dataset and network.} Following prior works~\citep{xu2022geodiff}, we utilize the GEOM-QM9~\citep{ramakrishnan2014quantum} and GEOM-Drugs~\citep{axelrod2022geom} datasets. 
The former dataset comprises small molecules of up to 9 heavy atoms, while the latter contains larger drug-like compounds. 
We reuse the data split provided by \citet{xu2022geodiff}. For both datasets, the training dataset comprises $40,000$ molecules, each with $5$ conformations, resulting in $200,000$ conformations in total.  
The test split includes $200$ distinctive molecules, with  $14,324$ conformations for Drugs and $22,408$ conformations for QM9.

Following \cite{xu2022geodiff}, we use an equivariant network GFN as the denoising network for conformation generation.
The detailed description of evaluation metrics and model architecture is in Appendix ~\ref{appsec:conformation_gene}.

\textbf{Conformation generation.} The comparison with \GeoDiff on the GEOM-QM9 dataset is reported in Table \ref{tab:qm9}. 
From the results, it is easy to see that \method significantly outperforms the \GeoDiff baseline on both metrics (COV-R and MAT-R) across different sampling steps. It indicates that by training with the substructure information, \method has a positive effect on the conformation generation task.
Moreover, \method with 500 steps achieves much better performance than \GeoDiff with 5000 steps on 5 out of 8 metrics, which implies our method can accelerate the sampling efficiency (10x).

\textbf{Domain generalization.}
To further illustrate the benefits of \method, we design two cross-domain tasks: (1) Training on QM9 (small molecular with up to 9 heavy atoms) and testing on Drugs (medium-sized organic compounds); (2) Training on Drugs and testing on QM9. The results (Table  \ref{tab:qm9_DG} and Appendix Table \ref{tab:Drugs_DG}) show that \method consistently outperforms \GeoDiff and other models trained on the in-domain dataset, demonstrating the introduced substructure effectively enhances the robustness and generalization of the diffusion model.

\section{Conclusion}
We present a novel diffusion model \method, which involves the subgraph constraint in the diffusion model by introducing a mask vector to the forward process. 
Benefiting from the expectation state and $k$-step same-subgraph diffusion, \method effectively boosts the perception of molecular substructure in the denoising network, thereby achieving state-of-the-art performance at various downstream property prediction tasks. There are several exciting avenues for future work. The mask distribution can be made flexible such that more chemical prior knowledge may be incorporated into efficient subgraph sampling. 
Besides, the proposed \method can be generalized to proteins such that the denoising network can learn meaningful secondary structures.


\bibliographystyle{icml2024.bst}
\bibliography{0_main.bib}

\newpage
\onecolumn
\appendix
\begin{center}
\Large
\textbf{Appendix}
 \\[20pt]
\end{center}

\etocdepthtag.toc{mtappendix}
\etocsettagdepth{mtchapter}{none}
\etocsettagdepth{mtappendix}{subsection}


\section{More Related Works}\label{sec:app_related_work}

\paragraph{Masks on diffusion models. }
Previous works also share a similar idea of subgraph (mask) diffusion, such as MDM~\citep{pan2023masked}, MDSM~\citep{lei2023masked}
and SSSD~\citep{alcaraz2022diffusion}. 
However, the difference between our \method and theirs mainly lies in the following two aspects: i) Usage: the mask matrix/vector in SSSD and MDSM is fixed in all training steps, which means some segments of the data (time series or images) will never be diffused. But our method samples the $\mbf s_t\sim p_{\s_t}{(\mcal S)}$ at each time step, hence a suitable discrete distribution $p(\mcal S)$ can ensure that almost all nodes can be added noise.
ii) Purpose: MDSM and MDM concentrate on self-supervised pre-training, while \method serves as a potent generative model  and self-supervised pre-training algorithm. Notably, when $\mbf s_t=\mbf s_0, \forall t$, \method can recover to MDSM. 
\vspace{-2mm}

\paragraph{Graph generation models.}
D3FG \citep{lin2023functionalgroupbased}: D3FG adopts three different diffusion models (D3PM, DDPM, and SO(3) Diffusion) to generate three different parts of molecules(linkerr types, center atom position, and functional group orientations), respectively. In general, these three parts can also be viewed as three subgraphs(subset). 
DiffPACK\citep{zhang2023diffpack} is an Autoregressive generative method that predicts the torsional angle $\chi_i (i=1,2,..,4)$ of protein side-chains with the condition $\chi_{1,...,i-1}$, where $\chi_i$ is a predefined subset of atoms. It uses a torsional-based diffusion model to approximate the distribution $p(\chi_i| \chi_{1,...,i-1})$, in which every subset $\chi_i$ needs a separate score network to estimate.  Essentially, both D3FG and DiffPACK can be viewed as selecting a subset first and then only adding noise on the fixed subset during the entire diffusion process.  In contrast, our method proposes to randomly sample a subset from mask distribution $p(S)$ in \textit{each time-step} during the forward process.
\cite{pmlr-v202-kong23b} proposes an autoregressive diffusion model named GraphARM, which absorbs one node in each time step by masking it along with its connecting edges during the forward process.   Differently from GraphARM, our \method selects a subgraph in each time step to inject the Gaussian noise, which is equivalent to masking several nodes during the forward process. In addition, the number of steps in GraphARM must be the same as the number of nodes due to the usage of the absorbing state, while our method can set any time-step during diffusion theoretically since we use the real-value Gaussian noise.
Concurrently, SubDiff~\cite{anonymous2024subdiff} is proposed to use subgraphs as minimum units to train a latent diffusion model, while our method directly involves the subgraph during the forward process, which is a new type of diffusion model.

\section{An important lemma for diffusion model}\label{appsec:lemma4.1}
\vspace{-2mm}
According to \citep{sohl2015deep, ho2020denoising}, the diffusion model is trained by optimizing the variational bound on the negative log-likelihood $-\log p_{\theta}(R^0)$, in which the tricky terms are $L_{t-1}=D_{KL}(q(R^{t-1}| R^t, R^0) || p_{\theta}(R^{t-1}| R^t)))$, $T\geq t>1$. 
Here we provide a lemma that tells us the posterior distribution $q(R^{t-1}| R^t, R^0)$ used in the training and sampling algorithms of the diffusion model can be determined by $q(R^t|R^{t-1}, R^0)$, $q(R^{t-1}|R^0)$. Formally, we have
\begin{snugshade}
\vspace{-.2cm}

\begin{restatable}[]{lemma}{restalemmtwo}
    \label{Lemma:diffusion}
    Assume the forward and reverse processes of the diffusion model are both Markov chains.
Given the forward Gaussian distribution $q(R^t|R^{t-1},R^0) = \mcal N(R^t; \mu_1 R^{t-1},\sigma_1^2 \textbf{I})$, $q(R^{t-1}|R^0) = \mcal N(R^{t-1}; \mu_2R^0,\sigma_2^2 \textbf{I})$ and $\epsilon_0 \sim \mcal N(\mbf 0, \mbf I)$, the distribution $q(R^{t-1}|R^t,R^0)$ is 
\vspace{-1mm}
\begin{align*}
\scalemath{0.8}{q(R^{t-1}|R^t,R^0)\propto\mcal N(R^{t-1}; \frac{1}{\mu_1}(R^t - \frac{\sigma_1^2}{\sqrt{\mu_1^2\sigma_2^2 + \sigma_1^2}}\epsilon_0 ), \frac{\sigma_1^2 \sigma_2^2 }{\mu_1^2 \sigma_2^2 + \sigma_1^2}\textbf{I}) }.
\end{align*}

Parameterizing $p_\theta(R^{t-1}|R^t)$ in the reverse process as $\mcal N(R^{t-1}; \frac{1}{\mu_1}(R^t - \frac{\sigma_1^2}{\sqrt{\mu_1^2\sigma_2^2 + \sigma_1^2}}\epsilon_\theta(R^t,t) ), \frac{\sigma_1^2 \sigma_2^2 }{\mu_1^2 \sigma_2^2 + \sigma_1^2}\textbf{I})$ 
, the training objective of the DPM can be written as
\vspace{-2mm}
\begin{align*}
    \scalemath{0.84}{\mcal L(\theta) =  \mbb E_{t,R^0, 
    \epsilon}\Big[ \frac{\sigma_1^2}{2\mu_1^2\sigma_2^2} \|\epsilon-\epsilon_{\theta }(\mu_1\mu_2 R^0 + \sqrt{\mu_1^2\sigma_2^2 + \sigma_1^2}\epsilon, t)\|^2\Big]},
\end{align*}
and the sampling (reverse) process is 
\vspace{-3mm}
\begin{align}
\label{eq:sampling_diffusion}
    \scalemath{0.8}{R^{t-1} = \frac{1}{\mu_1}\left(R^t - \frac{\sigma_1^2}{\sqrt{\mu_1^2\sigma_2^2 + \sigma_1^2}}\epsilon_{\theta}(R^t,t)\right) +  \frac{\sigma_1 \sigma_2 }{\sqrt{\mu_1^2 \sigma_2^2 + \sigma_1^2}} z ,}
\end{align}
where $ z\sim \mcal N(\mbf 0, \mbf I)$.
\end{restatable}
\vspace{-.1cm}
\end{snugshade}
Once we get the variables $(\mu_1,\sigma_1,\mu_2,\sigma_2)$, we can directly obtain the training objective and sampling process via lemma \ref{Lemma:diffusion}, which will help the design of new diffusion models.

\vspace{-2mm}

\paragraph{\textbf{\revise{Proof:}}}

Given the forward Gaussian distribution $q(R^t|R^{t-1},R^0) = \mcal N(R^t; \mu_1 R^{t-1},\sigma_1^2 I)$ and $q(R^{t-1}|R^0) = \mcal N(R^{t-1}; \mu_2R^0,\sigma_2^2 I)$, we have
\begin{align}
    q(R^t|R^0) =q(R^t|R^{t-1},R^0)q(R^{t-1}|R^0)= \mcal N(R^t; \mu_1\mu_2 R^{0},(\sigma_1^2+ \mu_1^2\sigma_2^2) I) \label{eq:02t}
\end{align}

From the DDPM, we know training a diffusion model 
should optimize the ELBO of the data
\begin{align}
    \scalemath{0.80}{\log p(\R)}&
\geq \scalemath{0.90}{\mathbb{E}_{q(\R^{1:T}|\R^0)}\left[\log \frac{p(\R^{0:T})}{q(\R^{1:T}|\R^0)}\right]}\\
&= \scalemath{0.8}{\underbrace{\mathbb{E}_{q(\R^{1}|\R^0)}\left[\log p_{\bm{\theta}}(\R^0|\R^1)\right]}_\text{reconstruction term} - \underbrace{\infdiv{q(\R^T|\R^0)}{p(\R^T)}}_\text{prior matching term} 
- \sum_{t=2}^{T} \underbrace{\mathbb{E}_{q(\R^{t}|\R^0)}\left[\infdiv{q(\R^{t-1}|\R^t, \R^0)}{p_{\bm{\theta}}(\R^{t-1}|\R^t)}\right]}_\text{denoising matching term}} \label{eq:elbo}
\end{align}

To compute the KL divergence $\infdiv{q(\R^{t-1}|\R^t, \R^0)}{p_{\bm{\theta}}(\R^{t-1}|\R^t)}$, we first rewrite $q(\R^{t-1}|\R^{t},\R^0)$  by Bayes rule
\begin{align}
    q(R^{t-1}|R^{t},R^0) &= \frac{q(R^{t}|R^{t-1},R^0) q(R^{t-1}|R^{0})}{q(R^{t}|R^{0})}\\
    &= \scalemath{0.94}{\frac{\mathcal{N}(R^{t} ; \mu_1 R^{t-1}, \sigma_1^2\textbf{I})\mathcal{N}(R^{t-1} ; \mu_2R^0, \sigma_2^2 \textbf{I})}{\mathcal{N}(R^{t} ; \mu_1\mu_2R^0, (\sigma_1^2+ \mu_1^2\sigma_2^2)\textbf{I})}}\\
&\propto \scalemath{0.94}{\text{exp}\left\{-\left[\frac{(R^{t} - \mu_1 R^{t-1})^2}{2\sigma_1^2} + \frac{(R^{t-1} - \mu_2 R^0)^2}{2\sigma_2^2} - \frac{(R^{t} - \mu_1\mu_2 R^{0})^2}{2(\sigma_1^2+ \mu_1^2\sigma_2^2)} \right]\right\}}\\
&= \scalemath{0.94}{\text{exp}\left\{-\frac{1}{2}\left[\frac{(R^{t} - \mu_1 R^{t-1})^2}{\sigma_1^2} + \frac{(R^{t-1} - \mu_2 R^0)^2}{\sigma_2^2} - \frac{(R^{t} - \mu_1\mu_2 R^{0})^2}{\sigma_1^2+ \mu_1^2\sigma_2^2} \right]\right\}}\\
&= \scalemath{0.94}{\text{exp}\left\{-\frac{1}{2}\left[\frac{(-2\mu_1 R^{t}R^{t-1} + \mu_1^2 (R^{t-1})^2)}{\sigma_1^2} + \frac{((R^{t-1})^2 - 2\mu_2R^{t-1} R^0)}{\sigma_2^2} + C(R^t, R^0)\right]\right\}} \label{eq:73}\\
&\propto \scalemath{0.94}{\text{exp}\left\{-\frac{1}{2}\left[- \frac{2\mu_1 R^{t}R^{t-1}}{\sigma_1^2} + \frac{\mu_1^2 (R^{t-1})^2}{\sigma_1^2} + \frac{(R^{t-1})^2}{\sigma_2^2} - \frac{2\mu_2R^{t-1} R^0}{\sigma_2^2}\right]\right\}}\\
&= \scalemath{0.94}{\text{exp}\left\{-\frac{1}{2}\left[(\frac{\mu_1^2}{\sigma_1^2} + \frac{1}{\sigma_2^2})(R^{t-1})^2 - 2\left(\frac{\mu_1R^{t}}{\sigma_1^2} + \frac{\mu_2R^0}{\sigma_2^2}\right)R^{t-1}\right]\right\}}\\
&= \scalemath{0.94}{\text{exp}\left\{-\frac{1}{2}\left[\frac{\sigma_1^2+ \mu_1^2\sigma_2^2}{\sigma_1^2\sigma_2^2}(R^{t-1})^2 - 2\left(\frac{\mu_1R^{t}}{\sigma_1^2} + \frac{\mu_2R^0}{\sigma_2^2}\right)R^{t-1}\right]\right\}}\\
&= \scalemath{0.94}{\text{exp}\left\{-\frac{1}{2}\left(\frac{\sigma_1^2+ \mu_1^2\sigma_2^2}{\sigma_1^2\sigma_2^2}\right)\left[(R^{t-1})^2 - 2\frac{\left(\frac{\mu_1R^{t}}{\sigma_1^2} + \frac{\mu_2R^0}{\sigma_2^2}\right)}{\frac{\sigma_1^2+ \mu_1^2\sigma_2^2}{\sigma_1^2\sigma_2^2}}R^{t-1}\right]\right\}}\\
&= \scalemath{0.94}{\text{exp}\left\{-\frac{1}{2}\left(\frac{\sigma_1^2+ \mu_1^2\sigma_2^2}{\sigma_1^2\sigma_2^2}\right)\left[(R^{t-1})^2 - 2\frac{\left(\frac{\mu_1R^{t}}{\sigma_1^2} + \frac{\mu_2R^0}{\sigma_2^2}\right)\sigma_1^2\sigma_2^2}{\sigma_1^2+ \mu_1^2\sigma_2^2}R^{t-1}\right]\right\}}\\
&= \scalemath{0.94}{\text{exp}\left\{-\frac{1}{2}\left(\frac{1}{\frac{\sigma_1^2\sigma_2^2}{\sigma_1^2+ \mu_1^2\sigma_2^2}}\right)\left[(R^{t-1})^2 - 2\frac{\mu_1\sigma_2^2R^{t} + \mu_2\sigma_1^2R^0}{\sigma_1^2+ \mu_1^2\sigma_2^2}R^{t-1}\right]\right\}}\\
&\propto \scalemath{0.94}{\mathcal{N}(R^{t-1} ;} \underbrace{\scalemath{0.94}{\frac{\mu_1\sigma_2^2R^{t} + \mu_2\sigma_1^2R^0}{\sigma_1^2+ \mu_1^2\sigma_2^2}}}_{\mu_q(R^t, R^0)}, \underbrace{\scalemath{0.94}{\frac{\sigma_1^2\sigma_2^2}{\sigma_1^2+ \mu_1^2\sigma_2^2}\textbf{I}}}_{\bm{\Sigma}_q(t)}) \label{eq:78}
\end{align}  

We can rewrite our variance equation as $\bm{\Sigma}_q(t) = \sigma_q^2(t)\textbf{I}$, where:
\begin{align}
    \sigma_q^2(t) =\frac{\sigma_1^2\sigma_2^2}{\sigma_1^2+ \mu_1^2\sigma_2^2}\label{eq:79}
\end{align}

From \eqref{eq:02t}, we have the relationship between $R^t$ and $R^0$:
\begin{align}
    R^0=\frac{R^t-\sqrt{\sigma_1^2+ \mu_1^2\sigma_2^2}\epsilon}{\mu_1\mu_2}
\end{align}
Substituting this into $\mu_q(R^t, R^0)$, we can get
\begin{align}
   \mu_q(R^t, R^0) &=  \frac{\mu_1\sigma_2^2R^{t} + \mu_2\sigma_1^2R^0}{\sigma_1^2+ \mu_1^2\sigma_2^2}\\
   &=\frac{\mu_1\sigma_2^2R^{t} + \mu_2\sigma_1^2 \frac{R^t-\sqrt{\sigma_1^2+ \mu_1^2\sigma_2^2}\epsilon}{\mu_1\mu_2}}{\sigma_1^2+ \mu_1^2\sigma_2^2}\\
   &=\frac{\mu_1\sigma_2^2R^{t} + \frac{\sigma_1^2 R^2}{\mu_1}-\frac{\sigma_1^2\sqrt{\sigma_1^2+ \mu_1^2\sigma_2^2}\epsilon}{\mu_1}}{\sigma_1^2+ \mu_1^2\sigma_2^2}\\
   &=\frac{1}{\mu_1}R^t - \frac{\sigma_1^2}{\mu_1\sqrt{\sigma_1^2+ \mu_1^2\sigma_2^2}}\epsilon
\end{align}

Thus,
\begin{align}
      q(R^{t-1}|R^{t},R^0) \propto \scalemath{0.94}{\mathcal{N}(R^{t-1} ;} \underbrace{\scalemath{0.94}{\frac{1}{\mu_1}(R^t - \frac{\sigma_1^2}{\sqrt{\sigma_1^2+ \mu_1^2\sigma_2^2}}\epsilon)}}_{\mu_q(R^t, t)}, \underbrace{\scalemath{0.94}{\frac{\sigma_1^2\sigma_2^2}{\sigma_1^2+ \mu_1^2\sigma_2^2}\textbf{I}}}_{\bm{\Sigma}_q(t)})
\end{align}
Parameterizing $p_\theta(R^{t-1}|R^t)$ in the reverse process as $\mcal N(R^{t-1}; \frac{1}{\mu_1}(R^t - \frac{\sigma_1^2}{\sqrt{\mu_1^2\sigma_2^2 + \sigma_1^2}}\epsilon_\theta(R^t,t) ), \frac{\sigma_1^2 \sigma_2^2 }{\mu_1^2 \sigma_2^2 + \sigma_1^2}\textbf{I})$ 
, and the corresponding optimization problem becomes:
\begin{align}
& \quad \,\argmin_{{\theta}} \infdiv{q(R^{t-1}|R^t, R^0)}{p_{{\theta}}(R^{t-1}|R^t)} \nonumber \\
&= \argmin_{{\theta}}\infdiv{\mathcal{N}\left(R^{t-1}; {\mu}_q,{\Sigma}_q\left(t\right)\right)}{\mathcal{N}\left(R^{t-1}; {\mu}_{{\theta}},{\Sigma}_q\left(t\right)\right)}\\
&=\argmin_{{\theta}}\frac{1}{2\sigma_q^2(t)}\left[\left\lVert\frac{\sigma_1^2}{\mu_1\sqrt{\sigma_1^2+ \mu_1^2\sigma_2^2}}\epsilon_0 - \frac{\sigma_1^2}{\mu_1\sqrt{\sigma_1^2+ \mu_1^2\sigma_2^2}}\epsilon_{{\theta}}(R^t, t)\right\rVert_2^2\right]\\
&=\argmin_{{\theta}}\frac{1}{2\sigma_q^2(t)}\left[\left\lVert \frac{\sigma_1^2}{\mu_1\sqrt{\sigma_1^2+ \mu_1^2\sigma_2^2}}({\epsilon}_0 - {\hat\epsilon}_{{\theta}}(R^t, t))\right\rVert_2^2\right]\\
&=\argmin_{{\theta}}\frac{1}{2\sigma_q^2(t)}\left(\frac{\sigma_1^2}{\mu_1\sqrt{\sigma_1^2+ \mu_1^2\sigma_2^2}}\right)^2\left[\left\lVert{\epsilon}_0 - {\hat\epsilon}_{{\theta}}(R^t, t)\right\rVert_2^2\right] \\
&=\argmin_{{\theta}}\frac{\sigma_1^2}{2\sigma_2^2\mu_1^2}\left[\left\lVert{\epsilon}_0 - {\hat\epsilon}_{{\theta}}(R^t, t)\right\rVert_2^2\right] 
\end{align}
Therefore, the training objective of the DPM can be written as
\begin{align}
    \mcal L(\theta) =  \mbb E_{t,R^0, 
    \epsilon}[ \frac{\sigma_1^2}{2\mu_1^2\sigma_2^2} \|\epsilon-\epsilon_{\theta }(\mu_1\mu_2 R^0 + \sqrt{\mu_1^2\sigma_2^2 + \sigma_1^2}\epsilon, t)\|^2],
\end{align}

During the reverse process, we sample $R^{t-1} \sim p_{\theta}(R^{t-1}|R^t)$.
Formally, the sampling (reverse) process is 
\begin{align}
\label{eq:app_sampling_diffusion}
    R^{t-1} =
    \frac{1}{\mu_1}\left(R^t - \frac{\sigma_1^2}{\sqrt{\mu_1^2\sigma_2^2 + \sigma_1^2}}\epsilon_{\theta}(R^t,t)\right) +  \frac{\sigma_1 \sigma_2 }{\sqrt{\mu_1^2 \sigma_2^2 + \sigma_1^2}} z , \quad z\sim \mcal N(\mbf 0, \mbf I)
\end{align}

\section{ Derivations of Training Objectives}
\subsection{\method (1-same step and without expectation state)}\label{sec:app_c1}

Here, we utilize the binary characteristic of the mask vector to derive the ELBO  for \method, and we also provide a general proof in sec. \ref{sec:app_ELOD}:
\begingroup
\allowdisplaybreaks
\begin{align}
\scalemath{0.90}{\log p(R^0)}
&\geq \scalemath{0.90}{\mathbb{E}_{q(R^{1:T}, s_{1:T}|R^0)}\left[\log \frac{p(R^{0:T}, s_{1:T})}{q(R^{1:T}|R^0,s_{1:T})q(s_{1:T})}\right]}\\
&= \scalemath{0.90}{\mathbb{E}_{q(R^{1:T}, s_{1:T}|R^0)}\left[\log \frac{p(R^T)\prod_{t=1}^{T}p_{\bm{\theta}}(R^{t-1}, s_{t}|R^t)}{\prod_{t = 1}^{T}q(
R^{t}|R^{t-1},s_t) q(s_t )}\right]}\\
&= \scalemath{0.90}{\mathbb{E}_{q(R^{1:T}, s_{1:T}|R^0)}\left[\log \frac{p(R^T)\prod_{t=1}^{T}p_{\bm{\theta}}(R^{t-1}|R^t) p_\theta(s_{t}|R^t)}{\prod_{t = 1}^{T}q(
R^{t}|R^{t-1},s_t) q(s_t)}\right]}\\
&= \scalemath{0.90}{\mathbb{E}_{q(R^{1:T}, s_{1:T}|R^0)}\left[\log \frac{\prod_{t=1}^{T} p_\theta(s_{t}|R^t)}{\prod_{t = 1}^{T} q(s_t)} + \log \frac{p(R^T)\prod_{t=1}^{T}p_{\bm{\theta}}(R^{t-1}|R^t)}{\prod_{t = 1}^{T}q(
R^{t}|R^{t-1},s_t)}\right]}\\
&= \scalemath{0.90}{\underbrace{\mathbb{E}_{q(R^{1:T}, s_{1:T}|R^0)}\left[\sum_{t = 1}^{T}\log \frac{p_\theta(s_{t}|R^t)}{ q(s_t)}\right]}_\text{mask prediction term} + \mathbb{E}_{q(R^{1:T}, s_{1:T}|R^0)}\left[\log \frac{p(R^T)\prod_{t=1}^{T}p_{\bm{\theta}}(R^{t-1}|R^t)}{\prod_{t = 1}^{T}q(
R^{t}|R^{t-1},s_t)}\right]}\\
\end{align}
\endgroup
The first term is mask prediction while the second term is similar to the ELBO of the classical diffusion model. The only difference is the $s_t$ in $q(R^t|R^{t-1},s_t)$. 
 According to Bayes rule, we can rewrite each transition as: 
\begin{align}
q(R^t | R^{t-1}, R^0, s_t) =
 \begin{cases} 1
   \frac{q(R^{t-1}|R^t,   R^0)q(R^t|R^0)}{q(R^{t-1}|R^0)},  &  \text{ if } s_{t}=1 \\
  \delta_{R_{t-1}}(R_t). & \text{ if } s_{t} =0
\end{cases}
\end{align}
where $\delta_a(x):=\delta(x-a)$ is Dirac delta function, that is, $\delta_a(x)=0$ if $x\neq a$ and $\int_{-\infty }^{\infty}\delta_a(x)dx=1$. Without loss of generality, assume that $s_1$ and $s_T$ both equal $1$.
Armed with this new equation, we drive the second term:
\begingroup
\allowdisplaybreaks
\begin{align}
& \mathbb{E}_{q(R^{1:T}, s_{1:T}|R^0)} \left[\log \frac{p(R^T)\prod_{t=1}^{T}p_{\bm{\theta}}(R^{t-1}|R^t)}{\prod_{t = 1}^{T}q(
R^{t}|R^{t-1},s_t)}\right] \\
&= \scalemath{0.90}{\mathbb{E}_{q(R^{1:T}, s_{1:T}|R^0)}\left[\log \frac{p(R^T)p_{\bm{\theta}}(R^0|R^1)\prod_{t=2}^{T}p_{\bm{\theta}}(R^{t-1}|R^t)}{q(R^1|R^0)\prod_{t = 2}^{T}q(R^{t}|R^{t-1},s_t)}\right]}\\
&= \scalemath{0.90}{\mathbb{E}_{q(R^{1:T}, s_{1:T}|R^0)}\left[\log \frac{p(R^T)p_{\bm{\theta}}(R^0|R^1)\prod_{t=2}^{T}p_{\bm{\theta}}(R^{t-1}|R^t)}{q(R^1|R^0)\prod_{t = 2}^{T}q(R^{t}|R^{t-1}, R^0,s_t)}\right]}\\
&= \scalemath{0.90}{\mathbb{E}_{q(R^{1:T}, s_{1:T}|R^0)}\left[\log \frac{p_{\bm{\theta}}(R^T)p_{\bm{\theta}}(R^0|R^1)}{q(R^1|R^0)} + \log \prod_{t=2}^{T}\frac{p_{\bm{\theta}}(R^{t-1}|R^t)}{q(R^{t}|R^{t-1}, R^0,s_t)}\right]}\\
&= \scalemath{0.90}{\mathbb{E}_{q(R^{1:T}, s_{1:T}|R^0)}\left[\log \frac{p(R^T)p_{\bm{\theta}}(R^0|R^1)}{q(R^1|R^0)} + \log \prod_{t\in\{t|s_t=1\}}\frac{p_{\bm{\theta}}(R^{t-1}|R^t)}{\frac{q(R^{t-1}|R^{t}, R^0)q(R^t|R^0)}{q(R^{t-1}|R^0, s_1)}} + \log \prod_{t\in\{t|s_t=0\}}\frac{p_{\bm{\theta}}(R^{t-1}|R^t)}{\delta_{R^{t-1}}(R^{t})} \right]}\\
&= \scalemath{0.90}{\mathbb{E}_{q(R^{1:T}|R^0)}\left[\log \frac{p(R^T)p_{\bm{\theta}}(R^0|R^1)}{q(R^1|R^0)} + \log \prod_{t\in\{t|s_t=0\}}\frac{p_{\bm{\theta}}(R^{t-1}|R^t)}{\delta_{R^{t-1}}(R^{t})} + \log \prod_{t\in\{t|s_t=1\}} \frac{p_{\bm{\theta}}(R^{t-1}|R^t)}{\frac{q(R^{t-1}|R^{t}, R^0)\cancel{q(R^t|R^0)}}{\cancel{q(R^{t-1}|R^0)}}}\right]}\\   
&= \scalemath{0.90}{\mathbb{E}_{q(R^{1:T}|R^0)}\left[ \log \prod_{t\in\{t|s_t=0\}}\frac{p_{\bm{\theta}}(R^{t-1}|R^t)}{\delta_{R^{t-1}}(R^{t})} +\log \frac{p(R^T)p_{\bm{\theta}}(R^0|R^1)}{\cancel{q(R^1|R^0)}} + \log \frac{\cancel{q(R^1|R^0)}}{q(R^T|R^0)} + \log \prod_{t\in\{t|s_t=1\}}\frac{p_{\bm{\theta}}(R^{t-1}|R^t)}{q(R^{t-1}|R^{t}, R^0)}\right]}\\
&= \scalemath{0.90}{\mathbb{E}_{q(R^{1:T}|R^0)}\left[\sum_{t\in\{t|s_t=0\}}\log \frac{p_{\bm{\theta}}(R^{t-1}|R^t)}{\delta_{R^{t-1}}(R^{t})} +\log \frac{p(R^T)p_{\bm{\theta}}(R^0|R^1)}{q(R^T|R^0)} +  \sum_{t\in\{t|s_t=1\}}\log\frac{p_{\bm{\theta}}(R^{t-1}|R^t)}{q(R^{t-1}|R^{t}, R^0)}\right]}\\
&= \scalemath{0.90}{\sum_{t\in\{t|s_t=0\}}\mathbb{E}_{q(R^{1:T}|R^0)}\left[\log \frac{p_{\bm{\theta}}(R^{t-1}|R^t)}{\delta_{R^{t-1}}(R^{t})}\right] + \mathbb{E}_{q(R^{1:T}|R^0)}\left[\log p_{\bm{\theta}}(R^0|R^1)\right]  } 
\\&\scalemath{0.90}{\quad +\mathbb{E}_{q(R^{1:T}|R^0)}\left[\log \frac{p(R^T)}{q(R^T|R^0)}\right] + \sum_{t\in\{t|s_t=1\}}\mathbb{E}_{q(R^{1:T}|R^0)}\left[\log\frac{p_{\bm{\theta}}(R^{t-1}|R^t)}{q(R^{t-1}|R^{t}, R^0)}\right]}\\
&= \scalemath{0.90}{\sum_{t\in\{t|s_t=0\}}\mathbb{E}_{q(R^{1:T}|R^0)}\left[\log \frac{p_{\bm{\theta}}(R^{t-1}|R^t)}{\delta_{R^{t-1}}(R^{t})}\right] +\mathbb{E}_{q(R^{1}|R^0)}\left[\log p_{\bm{\theta}}(R^0|R^1)\right] } \\
& \quad \scalemath{0.90}{+ \mathbb{E}_{q(R^{T}|R^0)}\left[\log \frac{p(R^T)}{q(R^T|R^0)}\right] + \sum_{t\in\{t|s_t=1\}}\mathbb{E}_{q(R^{t}, R^{t-1}|R^0)}\left[\log\frac{p_{\bm{\theta}}(R^{t-1}|R^t)}{q(R^{t-1}|R^{t}, R^0)}\right]}\\
&= \scalemath{0.9}{\underbrace{\sum_{t\in\{t|s_t=0\}}\mathbb{E}_{q(R^{1:T}|R^0)}\left[\log \frac{p_{\bm{\theta}}(R^{t-1}|R^t)}{\delta_{R^{t-1}}(R^{t})}\right]}_{\text{\textbf{decay term}}} +\underbrace{\mathbb{E}_{q(R^{1}|R^0)}\left[\log p_{\bm{\theta}}(R^0|R^1)\right]}_\text{reconstruction term} }
\\
&\quad \scalemath{0.9}{ - \underbrace{\infdiv{q(R^T|R^0)}{p(R^T)}}_\text{prior matching term} - \sum_{t\in\{t|s_t=1\}} \underbrace{\mathbb{E}_{q(R^{t}|R^0)}\left[\infdiv{q(R^{t-1}|R^t, R^0)}{p_{\bm{\theta}}(R^{t-1}|R^t)}\right]}_\text{denoising matching term}} \label{eq:51}
\end{align}
\endgroup

Here, the \textit{decay term} represents the terms with $s_t=0$, which are unnecessary to minimize when we set $p_{\bm\theta}(R^{t-1}|R^t):=\delta_{R^{t-1}}(R^{t})$.
Eventually, the ELOB can be rewritten as follows:
\begin{align}
\scalemath{0.90}{\log p(R^0)}
&\geq     \scalemath{0.90}{\sum_{t = 1}^{T}\underbrace{\mathbb{E}_{q(R^{1:T}|R^0)}\left[\log \frac{p_\vartheta(s_{t}|R^t)}{ q(s_t)}\right]}_\text{mask prediction term} +\underbrace{\mathbb{E}_{q(R^{1}|R^0)}\left[\log p_{\bm{\theta}}(R^0|R^1)\right]}_\text{reconstruction term} }
\notag\\
&\quad \scalemath{0.9}{ - \underbrace{\infdiv{q(R^T|R^0)}{p(R^T)}}_\text{prior matching term} - \sum_{t\in\{t|s_t=1\}} \underbrace{\mathbb{E}_{q(R^{t}|R^0)}\left[\infdiv{q(R^{t-1}|R^t, R^0)}{p_{\bm{\theta}}(R^{t-1}|R^t)}\right]}_\text{denoising matching term}} \label{eq:maskdiff_elbo}
\end{align}

The mask prediction term can be implemented by a node classifier and the denoising matching term can be calculated via Lemma ~\ref{Lemma:diffusion}. In detail,
\begin{align}
q(R^{t}|R^{t-1}, R^0) &= \mcal{N}(R^{t-1}, \sqrt{1-\beta_t s_t} R^{t-1},(\beta_{t}s_{t})\mbf I),\\
    q(R^{t-1}|R^{0})&=\mcal{N}(R^{t-1}, \sqrt{\bar\gamma_{t-1} } R^{0},(1-\bar\gamma_{t-1})\mbf I).    
\end{align}
Thus, the training objective of \method is:
\begin{align}
    \scalemath{0.9}{ \mcal L(\theta,\vartheta) = \mbb E_{t,R^0, 
    \epsilon}\left[ \frac{s_t\beta_t}{2(1-s_t\beta_t)(1-\bar\gamma_{t-1})} \|\epsilon-\epsilon_{\theta }(\sqrt{\bar\gamma_t}R^0 + \sqrt{(1-\bar\gamma_t)}\epsilon, t,\mcal G)\|^2 + \lambda \mathrm{BCE}(\s_t, s_\vartheta(\mcal G, \R^t,t))\right]}
\end{align}

To recover the existing work, we omit the mask prediction term (i.e. Let $p_{\theta}(s_t|R^t):=q(s_t)$) of \method in the main text.

 \subsection{ELBO of \method}\label{sec:app_ELOD}

Here, we can derive the ELBO for \method:
\begingroup
\allowdisplaybreaks
\begin{align}
\scalemath{0.90}{\log p(R^0)} &= \scalemath{0.9}{\log \int \int p(R^{0:T},s_{1:T}) dR^{1:T}}ds_{1:T}\\
&= \scalemath{0.9}{\log \int \int \frac{p(R^{0:T},s_{1:T})q(R^{1:T},s_{1:T}|R^0)}{q(R^{1:T},s_{1:T}|R^0)} dR^{1:T}}ds_{1:T}\\
&= \scalemath{0.9}{\log \int \int\left[\frac{p(R^{0:T},s_{1:T})q(R^{1:T}|R^0,s_{1:T})q(s_{1:T})}{q(R^{1:T},s_{1:T}|R^0)}\right]dR^{1:T}ds_{1:T}}\\
&= \scalemath{0.9}{\log \mbb{E}_{q(s_{1:T})} \mathbb{E}_{q(R^{1:T}|R^0, s_{1:T})}\left[\frac{p(R^{0:T},s_{1:T}))}{q(R^{1:T},s_{1:T}|R^0)} \right]}\\
&\geq \scalemath{0.90}{ \mathbb{E}_{q(R^{1:T}|R^0, s_{1:T})} \left[\log \mbb{E}_{q(s_{1:T})}  \frac{p(R^{0:T}, s_{1:T})}{q(R^{1:T}|R^0,s_{1:T})q(s_{1:T})} \right]}\\
&\geq \scalemath{0.90}{\mathbb{E}_{q(R^{1:T}, s_{1:T}|R^0)}\left[\log \frac{p(R^T)\prod_{t=1}^{T}p_{\bm{\theta}}(R^{t-1}, s_{t}|R^t)}{\prod_{t = 1}^{T}q(
R^{t}|R^{t-1},s_t) q(s_t )}\right]}\\
&= \scalemath{0.90}{\mathbb{E}_{q(R^{1:T}, s_{1:T}|R^0)}\left[\log \frac{p(R^T)\prod_{t=1}^{T}p_{\bm{\theta}}(R^{t-1}|R^t) p_\theta(s_{t}|R^t)}{\prod_{t = 1}^{T}q(
R^{t}|R^{t-1},s_t) q(s_t)}\right]}\\
&= \scalemath{0.90}{\mathbb{E}_{q(R^{1:T}, s_{1:T}|R^0)}\left[\log \frac{\prod_{t=1}^{T} p_\theta(s_{t}|R^t)}{\prod_{t = 1}^{T} q(s_t)} + \log \frac{p(R^T)\prod_{t=1}^{T}p_{\bm{\theta}}(R^{t-1}|R^t, s_t)}{\prod_{t = 1}^{T}q(
R^{t}|R^{t-1},s_t)}\right]}\\
&= \scalemath{0.90}{\underbrace{\mathbb{E}_{q(R^{1:T}, s_{1:T}|R^0)}\left[\sum_{t = 1}^{T}\log \frac{p_\theta(s_{t}|R^t)}{ q(s_t)}\right]}_\text{mask prediction term} + \mathbb{E}_{q(R^{1:T}, s_{1:T}|R^0)}\left[\log \frac{p(R^T)\prod_{t=1}^{T}p_{\bm{\theta}}(R^{t-1}|R^t, s_t)}{\prod_{t = 1}^{T}q(
R^{t}|R^{t-1},s_t)}\right]}\\
\end{align}
\endgroup

 According to Bayes rule, we can rewrite each transition as: 
\begin{align}
q(R^t | R^{t-1}, R^0, s_{1:t}) =
   \frac{q(R^{t-1}|R^t,  R^0, s_{1:t})q(R^t|R^0, s_{1:t})}{q(R^{t-1}|R^0, s_{1:t-1})},  
\end{align}
Armed with this new equation, we drive the second term:
\begingroup
\allowdisplaybreaks
\begin{align}
& \mathbb{E}_{q(R^{1:T}, s_{1:T}|R^0)} \left[\log \frac{p(R^T)\prod_{t=1}^{T}p_{\bm{\theta}}(R^{t-1}|R^t, s_t)}{\prod_{t = 1}^{T}q(
R^{t}|R^{t-1},s_t)}\right] \\
&= \scalemath{0.90}{\mathbb{E}_{q(R^{1:T}, s_{1:T}|R^0)}\left[\log \frac{p(R^T)p_{\bm{\theta}}(R^0|R^1)\prod_{t=2}^{T}p_{\bm{\theta}}(R^{t-1}|R^t, s_t)}{q(R^1|R^0,s_1)\prod_{t = 2}^{T}q(R^{t}|R^{t-1},s_t)}\right]}\\
&= \scalemath{0.90}{\mathbb{E}_{q(R^{1:T}, s_{1:T}|R^0)}\left[\log \frac{p(R^T)p_{\bm{\theta}}(R^0|R^1)\prod_{t=2}^{T}p_{\bm{\theta}}(R^{t-1}|R^t, s_t)}{q(R^1|R^0,s_1)\prod_{t = 2}^{T}q(R^{t}|R^{t-1}, R^0,s_{1:t})}\right]}\\
&= \scalemath{0.90}{\mathbb{E}_{q(R^{1:T}, s_{1:T}|R^0)}\left[\log \frac{p_{\bm{\theta}}(R^T)p_{\bm{\theta}}(R^0|R^1)}{q(R^1|R^0,s_1)} + \log \prod_{t=2}^{T}\frac{p_{\bm{\theta}}(R^{t-1}|R^t, s_t)}{q(R^{t}|R^{t-1}, R^0,s_{1:t})}\right]}\\
&= \scalemath{0.90}{\mathbb{E}_{q(R^{1:T}, s_{1:T}|R^0)}\left[\log \frac{p(R^T)p_{\bm{\theta}}(R^0|R^1)}{q(R^1|R^0,s_1)} + \log \prod_{t=2}^{T}\frac{p_{\bm{\theta}}(R^{t-1}|R^t, s_t)}{\frac{q(R^{t-1}|R^{t}, R^0,s_{1:t})q(R^t|R^0,s_{1:t})}{q(R^{t-1}|R^0, s_{1:t-1})}}  \right]}\\
&= \scalemath{0.90}{\mathbb{E}_{q(R^{1:T},s_{1:t}|R^0)}\left[\log \frac{p(R^T)p_{\bm{\theta}}(R^0|R^1)}{q(R^1|R^0,s_1)} + \log \prod_{t=2}^{T} \frac{p_{\bm{\theta}}(R^{t-1}|R^t, s_t)}{\frac{q(R^{t-1}|R^{t}, R^0,s_{1:t})\cancel{q(R^t|R^0,s_{1:t})}}{\cancel{q(R^{t-1}|R^0,s_{1:t-1})}}}\right]}\\   
&= \scalemath{0.90}{\mathbb{E}_{q(R^{1:T},s_{1:t}|R^0)}\left[ \log \frac{p(R^T)p_{\bm{\theta}}(R^0|R^1)}{\cancel{q(R^1|R^0,s_1)}} + \log \frac{\cancel{q(R^1|R^0,s_1)}}{q(R^T|R^0,s_{1:T})} + \log \prod_{t=2}^{T}\frac{p_{\bm{\theta}}(R^{t-1}|R^t, s_t)}{q(R^{t-1}|R^{t}, R^0, s_{1:t})}\right]}\\
&= \scalemath{0.90}{\mathbb{E}_{q(R^{1:T}, s_{1:t}|R^0)}\left[\log \frac{p(R^T)p_{\bm{\theta}}(R^0|R^1)}{q(R^T|R^0,s_{1:T})} +  \sum_{t=2}^{T}\log\frac{p_{\bm{\theta}}(R^{t-1}|R^t, s_t)}{q(R^{t-1}|R^{t}, R^0, s_{1:t})}\right]}\\
&= \scalemath{0.90}{\mathbb{E}_{q(R^{1:T}, s_{1:t}|R^0)}\left[\log p_{\bm{\theta}}(R^0|R^1)\right]  } 
\\&\scalemath{0.90}{\quad +\mathbb{E}_{q(R^{1:T}, s_{1:t}|R^0)}\left[\log \frac{p(R^T)}{q(R^T|R^0,s_{1:T})}\right] + \sum_{t=2}^{T}\mathbb{E}_{q(R^{1:T}, s_{1:t}|R^0)}\left[\log\frac{p_{\bm{\theta}}(R^{t-1}|R^t, s_t)}{q(R^{t-1}|R^{t}, R^0, s_{1:t})}\right]}\\
&= \scalemath{0.90}{\mathbb{E}_{q(R^{1},s_1|R^0)}\left[\log p_{\bm{\theta}}(R^0|R^1)\right] } \\
& \quad \scalemath{0.90}{+ \mathbb{E}_{q(R^{T}|R^0,s_{1:T})q(s_{1:T})}\left[\log \frac{p(R^T)}{q(R^T|R^0,s_{1:T})}\right] + \sum_{t=2}^{T}\mathbb{E}_{q(R^{t}, R^{t-1}, s_{1:t}|R^0)}\left[\log\frac{p_{\bm{\theta}}(R^{t-1}|R^t, s_t)}{q(R^{t-1}|R^{t}, R^0, s_{1:t})}\right]}\\
&= \scalemath{0.9}{ \underbrace{\mathbb{E}_{q(R^1,s_1|R^0)}\left[\log p_{\bm{\theta}}(R^0|R^1)\right]}_\text{reconstruction term} }
\\
&\quad \scalemath{0.9}{ - \underbrace{\mathbb{E}_{q(s_{1:t})}\infdiv{q(R^T|R^0,s_{1:T})}{p(R^T)}}_\text{prior matching term} - \sum_{t=2}^{T} \underbrace{\mathbb{E}_{q(R^{t},s_{1:t}|R^0)}\left[\infdiv{q(R^{t-1}|R^t, R^0, s_{1:t})}{p_{\bm{\theta}}(R^{t-1}|R^t, s_t)}\right]}_\text{denoising matching term}} \label{eq:51_}
\end{align}
\endgroup

Eventually, the ELOB can be rewritten as follows:
\begin{align}
\scalemath{0.90}{\log p(R^0)}
&\geq  \scalemath{0.90}{\sum_{t = 1}^{T}\underbrace{\mathbb{E}_{q(R^{t}, s_{t}|R^0)}\left[\log \frac{p_{\vartheta}(s_{t}|R^t)}{ q(s_t)}\right]}_\text{mask prediction term}} + \scalemath{0.9}{ \underbrace{\mathbb{E}_{q(R^1,s_1|R^0)}\left[\log p_{\bm{\theta}}(R^0|R^1,s_1)\right]}_\text{reconstruction term} }
\\&\scalemath{0.9}{ - \underbrace{\mathbb{E}_{q(s_{1:t})}\infdiv{q(R^T|R^0,s_{1:T})}{p(R^T)}}_\text{prior matching term} - \sum_{t=2}^{T} \underbrace{\mathbb{E}_{q(R^{t},s_{1:t}|R^0)}\left[\infdiv{q(R^{t-1}|R^t, R^0, s_{1:t})}{p_{\bm{\theta}}(R^{t-1}|R^t,s_t)}\right]}_\text{denoising matching term}} \label{eq:elob_norm}
\end{align}

The mask prediction term can be implemented by a node classifier $s_{\vartheta}$. For the denoising matching term, by Bayes rule, the $q(R^{t-1} | R^{t}, R^0, s_{1:t})$ can be written as:
\begin{align}
q(R^{t-1} | R^{t}, R^0, s_{1:t}) =
   \frac{q(R^{t}|R^{t-1},  R^0, s_{1:t})q(R^{t-1}|R^0, s_{1:t-1})}{q(R^{t}|R^0, s_{1:t})},  \label{eq:bayes_st}
\end{align}
For the naive \method, we have
\begin{align}
q(R^{t}|R^{t-1}, R^0, s_{1:t}) &:= \mcal{N}(R^{t-1}, \sqrt{1-\beta_t s_t} R^{t-1},(\beta_{t}s_{t})\mbf I),\\
    q(R^{t-1}|R^{0}, s_{1:t-1})&:=\mcal{N}(R^{t-1}, \sqrt{\bar\gamma_{t-1} } R^{0},(1-\bar\gamma_{t-1})\mbf I).    
\end{align}
Then the denoising matching term can also be calculated via Lemma ~\ref{Lemma:diffusion} (let $q(R^{t}|R^{t-1}, R^0) :=q(R^{t}|R^{t-1}, R^0, s_{1:t}) $ ,  $q(R^{t-1}|R^{0}) := q(R^{t-1}|R^{0}, s_{1:t-1})$ and $p_\theta(R^{t-1}|R^t) = p_\theta(R^{t-1})$).
Thus, the training objective of \method is:
\begin{align}
    \scalemath{0.9}{ \mcal L(\theta,\vartheta) = \mbb E_{t,R^0, 
    \epsilon}\left[ \frac{s_t\beta_t}{2(1-s_t\beta_t)(1-\bar\gamma_{t-1})} \|\epsilon-\epsilon_{\theta }(\sqrt{\bar\gamma_t}R^0 + \sqrt{(1-\bar\gamma_t)}\epsilon, t,\mcal G)\|^2 + \lambda \mathrm{BCE}(\s_t, s_\vartheta(\mcal G, \R^t,t))\right]}
\end{align}

\subsubsection{Expectation of $s_{1:T}$} \label{subsec_app:expect_s}
The denoising matching term in ~\eqref{eq:elob_norm} can be calculated by only sampling $(R^t, s_t)$ instead of $(R^t, s_{1:t})$. Specifically, we substitute \eqref{eq:bayes_st} into the denoising matching term:
\begin{align}
    &\mathbb{E}_{q(R^{t}, R^{t-1}, s_{1:t}|R^0)}\left[\log\frac{p_{\bm{\theta}}(R^{t-1}|R^t, s_t)}{q(R^{t-1}|R^{t}, R^0, s_{1:t})}\right]\\
    &\scalemath{0.9}{=\mathbb{E}_{q(R^{t}, R^{t-1}, s_{1:t}|R^0)}\left[\log\frac{p_{\bm{\theta}}(R^{t-1}|R^t, s_t)}{ \frac{q(R^{t}|R^{t-1},  R^0, s_{1:t})q(R^{t-1}|R^0, s_{1:t-1})}{q(R^{t}|R^0, s_{1:t})}}\right]} \\
    &=\scalemath{0.9}{\mathbb{E}_{q(R^{t}, R^{t-1}, s_{1:t}|R^0)}\left[\log\frac{p_{\bm{\theta}}(R^{t-1}|R^t, s_t)}{ \frac{q(R^{t}|R^{t-1},  R^0, s_{t})}{q(R^{t}|R^0, s_{1:t})}} -\log q(R^{t-1}|R^0, s_{1:t-1})\right]} \\
      &\ge \scalemath{0.9}{\mathbb{E}_{q(R^{t}, R^{t-1}, |R^0,s_{1:t})}\left[ \mathbb{E}_{q(s_{1:t})} \log {p_{\bm{\theta}}(R^{t-1}|R^t, s_t) q(R^{t}|R^0, s_{1:t})}\right]} \\
      &\scalemath{0.9}{\quad-\mathbb{E}_{q(s_{t})}\left[\log \underbrace{\mathbb{E}_{q(s_{1:t-1})} q(R^{t-1}|R^0, s_{1:t-1})}_{:=q(\mathbb{E}_s R^{t-1}|R^0)} + \log \underbrace{\mathbb{E}_{q(s_{1:t-1})} q(R^{t}|R^{t-1},  R^0, s_{1:t})}_{:=q(R^{t}|\mathbb{E}_s R^{t-1},  R^0, s_{t})}\right]}\\
      &
      =\scalemath{0.8}{\mathbb{E}_{q(R^{t}, R^{t-1}, |R^0,s_{1:t})}\left[ \mathbb{E}_{q(s_{1:t})} \log\frac{p_{\bm{\theta}}(R^{t-1}|R^t, s_t)}{ \frac{1}{q(R^{t}|R^0, s_{1:t})}} -\mathbb{E}_{q(s_{t})}\log q(\mathbb{E}_s R^{t-1}|R^0) - \mathbb{E}_{q(s_{t})}\log q(R^{t}|\mathbb{E}_s R^{t-1},  R^0, s_{t}) \right]}\\
      & = \scalemath{0.8}{\mathbb{E}_{q(R^{t}, R^{t-1}, |R^0,s_{1:t})}\left[ \mathbb{E}_{q(s_{1:t})} \log\frac{p_{\bm{\theta}}(R^{t-1}|R^t, s_t)}{ \frac{q(R^{t}|\mathbb{E}_s R^{t-1},  R^0, s_{t})q(\mathbb{E}_s R^{t-1}|R^0)}{q(R^{t}|R^0, s_{1:t})}}  \right]}\\
        & = \scalemath{0.9}{\underbrace{\mathbb{E}_{q(R^{t}, R^{t-1}, s_{1:t}|R^0)}\left[ \log\frac{p_{\bm{\theta}}(R^{t-1}|R^t, s_t)}{ \frac{q(R^{t}|\mathbb{E}_s R^{t-1},  R^0, s_{t})q(\mathbb{E}_s R^{t-1}|R^0)}{q(R^{t}|R^0, s_{1:t})}}  \right]}_\text{denoising matching term}}\\ 
        & = \scalemath{0.9}{{\mathbb{E}_{q(R^{t}, R^{t-1}, s_{1:t}|R^0)}\left[ \log\frac{p_{\bm{\theta}}(R^{t-1}|R^t, s_t)}{ \hat{q}(R^{t-1} | R^{t}, R^0, s_{1:t})}\right]}}\\ 
        &=\scalemath{0.9}{ \underbrace{\mathbb{E}_{q(R^{t},s_{1:t}|R^0)}\left[\infdiv{\hat{q}(R^{t-1}|R^t, R^0, s_{1:t})}{p_{\bm{\theta}}(R^{t-1}|R^t, s_t)}\right]}_\text{denoising matching term}} 
\end{align}
 Thus, we should focus on calculating the distribution 
 \begin{align}    
 \hat{q}(R^{t-1} |  R^{t}, R^0, s_{1:t}) :=
   \frac{q( R^{t}|\mbb E_s R^{t-1},  R^0, s_{t})q(\mbb E_s R^{t-1}|R^0)}{q(R^{t}|R^0, s_{1:t})}
   \end{align}

By lemma ~\ref{Lemma:diffusion}, if we can gain the expression of $q( R^{t}|\mbb E_s R^{t-1},  R^0, s_{t})$ and $q(\mbb E_s R^{t-1}|R^0)$, we can get the training objective and sampling process. 
 \subsection{Single-step subgraph diffusion } 

 \subsubsection{Training}
 \paragraph{I: Step $0$ to Step $t-1$ ($R^0\to R^{t-1}$):}
 The state space of the mask diffusion should be the mean of the random state.
 
\begin{align}
     \mbb E_s R^t \sim \mcal N(\mbb E_s R^t ; \sqrt{1-\beta_t}\mbb E_s R^{t-1}, \beta_t I)
\end{align}
\begin{align}
\label{eq:t-step_transfer}
    q(R^{t}|R^{0},s_{1:t})=\mcal{N}(R^{t}, \sqrt{\bar\gamma_t } R^{0},(1-\bar\gamma_t)\mbf I).
\end{align}
 Form \eqref{eq:t-step_transfer}, we have:
 \begin{align}
    R^t &= \sqrt{1- s_t\beta_t } R^{t-1}   + \sqrt{s_t\beta_t} \epsilon_{t-1} \\
    \mbb E R^t &= (p\sqrt{1-\beta_t}+1-p)\mbb{E} R^{t-1} + p\sqrt{\beta_t}\epsilon_{t-1} \\
    &=(p\sqrt{1-\beta_t}+1-p)(p\sqrt{1-\beta_{t-1}}+1-p)\mbb{E} R^{t-2} +  (p\sqrt{1-\beta_t}+1-p)p\sqrt{\beta_{t-1}}\epsilon_{t-2}+ p\sqrt{\beta_t}\epsilon_{t-1} \\
    &=(p\sqrt{1-\beta_t}+1-p)(p\sqrt{1-\beta_{t-1}}+1-p)\mbb{E} R^{t-2} +  \sqrt{[(p\sqrt{1-\beta_t}+1-p)p\sqrt{\beta_{t-1}}]^2+ [p\sqrt{\beta_t}]^2} \epsilon_{t-2} \\
    &=....\\
    &= \prod_{i=1}^{t}(p\sqrt{1-\beta_i}+1-p) R^{0} + \sqrt{ [\prod_{j=2}^{t}(p\sqrt{1-\beta_j}+1-p)p\sqrt{\beta_1}]^2 + [\prod_{j=3}^{t}(p\sqrt{1-\beta_j}+1-p)p\sqrt{\beta_2}]^2 + ... +}\epsilon_{0} \\
    & =\prod_{i=1}^{t}(p\sqrt{1-\beta_i}+1-p) R^{0} + \sqrt{ \sum_{i=1}^{t} [\prod_{j=i+1}^{t}(p\sqrt{1-\beta_j}+1-p)p\sqrt{\beta_{i}}]^2}\\
    & = \prod_{i=1}^{t}\sqrt{\alpha_i} R^{0} + \sqrt{ \sum_{i=1}^{t} [\prod_{j=i+1}^{t}\sqrt{\alpha_i} 
    p\sqrt{\beta_{i}}]^2}\epsilon_{0}\\
    & = \prod_{i=1}^{t}\sqrt{\alpha_i} R^{0} + p \sqrt{ \sum_{i=1}^{t} \prod_{j=i+1}^{t}\alpha_j \beta_{i}}\epsilon_{0}\\
    & = \sqrt{\bar{\alpha}_t} R^{0} + p \sqrt{ \sum_{i=1}^{t} 
    \frac{\bar{\alpha}_t}{\bar{\alpha}_i} \beta_{i}}\epsilon_{0}\\
 \end{align}
 where $\alpha_i := (p\sqrt{1-\beta_i}+1-p)^2$ and $\bar{\alpha}_t = \prod_{i=1}^{t}\alpha_i$.

  \begin{align}
    q(\mbb E R^{t}|R^{0}) &=  \mcal{N}(R^{t}; \sqrt{\bar{\alpha}_t} R^{0},  p^2\sum_{i=1}^{t} 
    \frac{\bar{\alpha}_t}{\bar{\alpha}_i} \beta_{i} I)
\end{align}

 \paragraph{II: Step $t-1$ to Step $t$ ($R^{t-1}\to R^{t}$):}
 We build the step $t-1 \to t$ is a discrete transition from  $q(\R^{t-1}|\R^{0})$, with 
 \begin{align}
    q(\mbb E_s R^{t-1}|R^{0}) &=  \mcal{N}(R^{t-1}; \prod_{i=1}^{t-1}\sqrt{\alpha_i} R^{0},  p^2\sum_{i=1}^{t-1} \prod_{j=i+1}^{t-1}\alpha_j \beta_{i} I)
    \\
     q(R^t|\mbb E_s R^{t-1}, s_t ) &=\mcal N(R^t ; \sqrt{1-s_t\beta_t}\mbb E R^{t-1},s_t \beta_t I)
\end{align}

\begin{align}
    R^{t} &=  \sqrt{1- s_t\beta_t } \mbb E R^{t-1} + \sqrt{s_t\beta_t} \epsilon_{t-1} \\
    &=\sqrt{1- s_t\beta_t }\left(\sqrt{\bar{\alpha}_{t-1}} R^{0} + p \sqrt{ \sum_{i=1}^{t-1} 
    \frac{\bar{\alpha}_{t-1}}{\bar{\alpha}_i} \beta_{i}}\epsilon_{0}\right) + \sqrt{s_t\beta_t} \epsilon_{t-1} \\
    &=\sqrt{1- s_t\beta_t }\sqrt{\bar{\alpha}_{t-1}} R^{0} + p\sqrt{1- s_t\beta_t } \sqrt{ \sum_{i=1}^{t-1} 
    \frac{\bar{\alpha}_{t-1}}{\bar{\alpha}_i} \beta_{i}}\epsilon_{0} + \sqrt{s_t\beta_t} \epsilon_{t-1}  \\
    &=\sqrt{1- s_t\beta_t }\sqrt{\bar{\alpha}_{t-1}} R^{0} + \sqrt{p^2(1- s_t\beta_t)  \sum_{i=1}^{t-1} 
    \frac{\bar{\alpha}_{t-1}}{\bar{\alpha}_i} \beta_{i} + s_t\beta_t }\epsilon_{0}
\end{align}

 \paragraph{ Step $0$ to Step $t$ ($R^0\to R^{t}$):}
 \begin{align}
    q(R^t|R^{0} ) &= \int q(R^t|\mbb E R^{t-1} ) q(\mbb E R^{t-1}|R^{0}) d \mbb ER^{t-1} \\
    &= \mcal{N}(R^{t}; \sqrt{1- s_t\beta_t }\sqrt{\bar{\alpha}_i} R^{0}, ( p^2(1- s_t\beta_t)  \sum_{i=1}^{t-1} 
    \frac{\bar{\alpha}_{t-1}}{\bar{\alpha}_i} \beta_{i} + s_t\beta_t) I)
\end{align}

Thus, from subsection \ref{subsec_app:expect_s}, the \textbf{training objective} of 1-step \method is:
\begin{align}
    \mcal L_{simple}(\theta, \vartheta ) = \mbb E_{t,R^0,s_t, 
    \epsilon}[s_t\|\epsilon-\epsilon_{\theta }(R^t, t)\|^2 - \mcal{BCE}(s_t, s_\vartheta(R^t,t))]
\end{align}
where $\mcal{BCE}(s_t, s_\vartheta) =s_t \log{s_\vartheta}(R^t,t) + (1-s_t) \log{(1-s_\vartheta}(R^t,t))$ is Binary Cross Entropy loss.
However, training the \method is not trivial. The challenges come from two aspects: 1) the mask predictor should be capable of perceiving the sensible noise change between $(t-1)$-th and $t$-th step. However, the noise scale $\beta_t$ is relatively small when $t$ is small, especially if the diffusion step is larger than a thousand, thereby mask predictor cannot precisely predict. 2) The accumulated noise for each node at $(t-1)$-th step would be mainly affected by the mask sampling from $1$ to $t-1$ step, which heavily increases the difficulty of predicting the noise added between $(t-1)$-step to $t$-step.

\subsubsection{Sampling}
Finally, the sampling can be written as:

\begin{align}
\label{eq:masked_sampling}
    R^{t-1} &= \frac{\left((1-s_t\beta_t)p^2\sum_{i=1}^{t-1} 
    \frac{\bar{\alpha}_{t-1}}{\bar{\alpha}_i} \beta_{i} + s_t\beta_t \right) R^t- \left( s_t\beta_t\sqrt{p^2 (1-s_t\beta_t)\sum_{i=1}^{t-1} 
    \frac{\bar{\alpha}_{t-1}}{\bar{\alpha}_i} \beta_{i} + s_t\beta_t}\right) \epsilon_\theta(R^t,t) }
    {\sqrt{1-s_t\beta_t}(s_t\beta_t + (1-s_t\beta_t)p^2\sum_{i=1}^{t-1} 
    \frac{\bar{\alpha}_{t-1}}{\bar{\alpha}_i} \beta_{i}   )}
    + \sigma_t z \\
    &=\frac{1}{\sqrt{1-s_t\beta_t}} R^t - \frac{ \left( s_t\beta_t\sqrt{p^2 (1-s_t\beta_t)\sum_{i=1}^{t-1} 
    \frac{\bar{\alpha}_{t-1}}{\bar{\alpha}_i} \beta_{i} + s_t\beta_t}\right)  }
    {\sqrt{1-s_t\beta_t}(s_t\beta_t + (1-s_t\beta_t)p^2\sum_{i=1}^{t-1} 
    \frac{\bar{\alpha}_{t-1}}{\bar{\alpha}_i} \beta_{i}   )}\epsilon_\theta(R^t,t)
    + \sigma_t z \\
        &=\frac{1}{\sqrt{1-s_t\beta_t}} R^t - \frac{  s_t\beta_t }
    {\sqrt{1-s_t\beta_t}\sqrt{s_t\beta_t + (1-s_t\beta_t)p^2\sum_{i=1}^{t-1} 
    \frac{\bar{\alpha}_{t-1}}{\bar{\alpha}_i} \beta_{i}   }}\epsilon_\theta(R^t,t)
    + \sigma_t z \\
\end{align}
where $s_t = s_\vartheta(R^t,t)$ and \begin{align}
    \sigma_t=\frac{s_\vartheta(R^t,t)\beta_t p^2 \sum_{i=1}^{t-1} 
    \frac{\bar{\alpha}_{t-1}}{\bar{\alpha}_i} \beta_{i}}
    {s_\vartheta(R^t,t)\beta_t + p^2(1-s_\vartheta(R^t,t)\beta_t)\sum_{i=1}^{t-1} 
    \frac{\bar{\alpha}_{t-1}}{\bar{\alpha}_i} \beta_{i}}
\end{align}

\section{Expectation state distribution}\label{sec:1_step_Mean}
 The state space of the mask diffusion should be the mean of the random state.
 
\begin{align}
     \mbb E_{s_t} R^t \sim \mcal N(\mbb E R^t ; \sqrt{1-\beta_t}\mbb E_{s_{t-1}} R^{t-1}, \beta_t I)
\end{align}
 Form \autoref{eq:t-step_transfer}, we have:
 \begin{align}
    R^t &= \sqrt{1- s_t\beta_t } R^{t-1}   + \sqrt{s_t\beta_t} \epsilon_{t-1} \\
    \mbb E R^t &= (p\sqrt{1-\beta_t}+1-p)\mbb{E} R^{t-1} + p\sqrt{\beta_t}\epsilon_{t-1} \\
    &=(p\sqrt{1-\beta_t}+1-p)(p\sqrt{1-\beta_{t-1}}+1-p)\mbb{E} R^{t-2} \\ 
    &
     \quad +  (p\sqrt{1-\beta_t}+1-p)p\sqrt{\beta_{t-1}}\epsilon_{t-2}+ p\sqrt{\beta_t}\epsilon_{t-1} \\
    &=(p\sqrt{1-\beta_t}+1-p)(p\sqrt{1-\beta_{t-1}}+1-p)\mbb{E} R^{t-2} \\ 
    &
     \quad +  \sqrt{[(p\sqrt{1-\beta_t}+1-p)p\sqrt{\beta_{t-1}}]^2+ [p\sqrt{\beta_t}]^2} \epsilon_{t-2} \\
    &=....\\
    &= \prod_{i=1}^{t}(p\sqrt{1-\beta_i}+1-p) R^{0} \\ 
    &
     \quad + \sqrt{ [\prod_{j=2}^{t}(p\sqrt{1-\beta_j}+1-p)p\sqrt{\beta_1}]^2 + [\prod_{j=3}^{t}(p\sqrt{1-\beta_j}+1-p)p\sqrt{\beta_2}]^2 + ... +}\epsilon_{0} \\
    & =\prod_{i=1}^{t}(p\sqrt{1-\beta_i}+1-p) R^{0} + \sqrt{ \sum_{i=1}^{t} [\prod_{j=i+1}^{t}(p\sqrt{1-\beta_j}+1-p)p\sqrt{\beta_{i}}]^2}\\
    & = \prod_{i=1}^{t}\sqrt{\alpha_i} R^{0} + \sqrt{ \sum_{i=1}^{t} [\prod_{j=i+1}^{t}\sqrt{\alpha_i} 
    p\sqrt{\beta_{i}}]^2}\epsilon_{0}\\
    & = \prod_{i=1}^{t}\sqrt{\alpha_i} R^{0} + p \sqrt{ \sum_{i=1}^{t} \prod_{j=i+1}^{t}\alpha_j \beta_{i}}\epsilon_{0}\\
    & = \sqrt{\bar{\alpha}_i} R^{0} + p \sqrt{ \sum_{i=1}^{t} 
    \frac{\bar{\alpha}_t}{\bar{\alpha}_i} \beta_{i}}\epsilon_{0}\\
 \end{align}
 where $\alpha_i := (p\sqrt{1-\beta_i}+1-p)^2$ and $\bar{\alpha}_t = \prod_{i=1}^{t}\alpha_i$.

Finally, the Expectation state distribution is:
  \begin{align}
    q(\mbb E R^t |R^{0}) &=  \mcal{N}(\mbb E R^{t}; \prod_{i=1}^{t}\sqrt{\alpha_i} R^{0},  p^2\sum_{i=1}^{t} \prod_{j=i+1}^{t}\alpha_j \beta_{i} I)
\end{align}

\begin{figure*}
    \centering
    \vspace{-0.2cm}
    \includegraphics[width=0.94\linewidth]{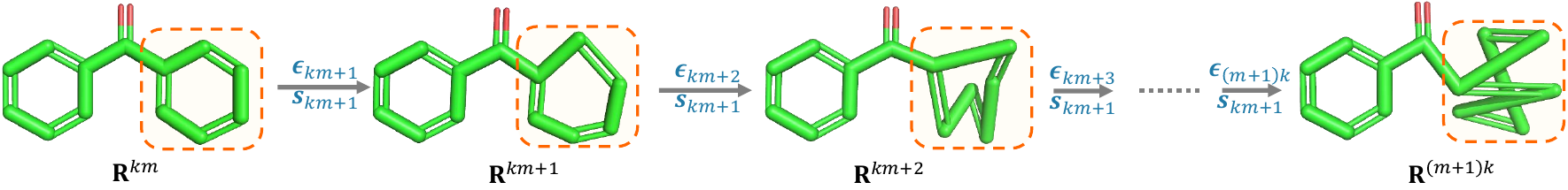}
     \caption{An example of $k$-step same subgraph diffusion, where the mask vectors are same as $\mbf s_{km+1}$ from step $km$ to $(m+1)k$, $m\in \mbb N^+$ .}
    \label{fig:fix_mask_diff}
        \vspace{-4mm}
\end{figure*}
\section{The derivation of \method}

When $t$ is an integer multiple of $k$,
\begin{align}
    \mbb ER^t &= \prod_{j=1}^{t/k}(p\sqrt{\prod_{i=(j-1)k+1}^{kj}(1-\beta_i)} + 1-p )R^0 \\
    &+\sqrt{\sum_{l=1}^{t/k}\left[\prod_{j=l+1}^{t/k}(p\sqrt{\prod_{i=(j-1)k+1}^{kj}(1-\beta_i)} + 1-p)p\sqrt{1-\prod_{i=(l-1)k+1}^{kl}(1-\beta_i)}\right]^2} \epsilon_0\\
    &=\prod_{j=1}^{t/k}\sqrt{\alpha_j} R^0 +  p\sqrt{\sum_{l=1}^{t/k} \prod_{j=l+1}^{t/k} \alpha_j (1-\prod_{i=(l-1)k+1}^{kl}(1-\beta_i))} \epsilon_0 \\
        &=\sqrt{\bar\alpha_{t/k}} R^0 +  p\sqrt{\sum_{l=1}^{t/k} \frac{\bar\alpha_{t/k}}{\bar\alpha_{l}} (1-\prod_{i=(l-1)k+1}^{kl}(1-\beta_i))} \epsilon_0
\end{align}
where $\alpha_j= (p\sqrt{\prod_{i=(j-1)k+1}^{kj}(1-\beta_i)} + 1-p)^2 $. 

When $t\in\mbb N$, we have
\begin{align}
    R^t &= \sqrt{\prod_{i=k \lfloor t/k\rfloor +1}^{t} (1-\beta_i s_{\lowbound{t/k}  })} \mbb E R^{\lowbound{t/k}\times k} + \sqrt{1- \prod_{i=k\lfloor t/k\rfloor +1}^{t} (1-\beta_i s_{\lowbound{t/k}  })} \epsilon_{\lowbound{t/k}\times k} \\
    &=\sqrt{\prod_{i=k\lfloor t/k\rfloor +1}^{t} (1-\beta_is_{\lowbound{t/k}  })} \left(\sqrt{\bar\alpha_{\lowbound{t/k}}} R^0 +  p\sqrt{\sum_{l=1}^{\lowbound{t/k}} \frac{\bar\alpha_{\lowbound{t/k}}}{\bar\alpha_{l}} (1-\prod_{i=(l-1)k+1}^{kl}(1-\beta_i))} \epsilon_0 \right)\\
    & \quad + \sqrt{1- \prod_{t=\lfloor t/k\rfloor}^{t} (1-\beta_is_{\lowbound{t/k} })} \epsilon_{\lowbound{t/k}} \\
    &= \sqrt{\prod_{i= k\lowbound{t/k}+1 }^t\gamma_i}\sqrt{\bar\alpha_{\lowbound{t/k}}} R^0 
     \\
    & \quad +\sqrt{ \left(\prod_{i= k\lowbound{t/k}+1  }^t \gamma_i\right) p^2 \sum_{l=1}^{\lowbound{t/k}}
    \frac{\bar\alpha_{\lowbound{t/k}}}{\bar\alpha_{l}} 
    (1-\prod_{i=(l-1)k+1}^{kl}(1-\beta_i)) + 
\left(1-\prod_{i= k\lowbound{t/k}+1  }^t\gamma_i\right) } \epsilon_0
\end{align}
where $\gamma_i = 1-\beta_is_{\lowbound{t/k}} $.

\begin{align}
    q(R^t|R^0) &=  \mcal{N}(R^{k\lowbound{t/k}}; \sqrt{\prod_{i= k\lowbound{t/k}+1 }^t\gamma_i}\sqrt{\bar\alpha_{\lowbound{t/k}}} R^0, \\  &\left(\left(\prod_{i= k\lowbound{t/k}+1  }^t \gamma_i\right) p^2 \sum_{l=1}^{\lowbound{t/k}}
    \frac{\bar\alpha_{\lowbound{t/k}}}{\bar\alpha_{l}} 
    (1-\prod_{i=(l-1)k+1}^{kl}(1-\beta_i)) + 
1-\prod_{i= k\lowbound{t/k}+1  }^t\gamma_i\right)  I) 
\end{align}
Let $m=\lowbound{t/k}$ , $\bar\gamma_i = \prod_{t=1}^i \gamma_t$, and $\bar\beta_t = \prod_{i=1}^{t}(1-\beta_i)$
\begin{align}
q(R^t|R^0) &=\mcal{N}(R^{km}; \sqrt{\frac{\bar\gamma_t}{\bar\gamma_{km}}}\sqrt{\bar\alpha_{m}} R^0, 
\left(\frac{\bar\gamma_t}{\bar\gamma_{km}} p^2 \sum_{l=1}^{m}
    \frac{\bar\alpha_{m}}{\bar\alpha_{l}} 
    (1-\frac{\bar\beta_{kl}}{\bar\beta_{(l-1)k}}) + 
1-\frac{\bar\gamma_t}{\bar\gamma_{km}}\right)  I) 
\end{align}

\subsubsection{Sampling}
\begin{align}
    \mu_1 &= \sqrt{ 1-s_{km+1}\beta_t},\\
    \sigma_1^2 &= s_{km+1}\beta_t \\
    \mu_2&= \sqrt{\frac{\bar\gamma_{t-1}}{\bar\gamma_{km}}}\sqrt{\bar\alpha_{m}} \\
    \sigma_2^2 &= \frac{\bar\gamma_{t-1}}{\bar\gamma_{km}} p^2 \sum_{l=1}^{m}
    \frac{\bar\alpha_{m}}{\bar\alpha_{l}} 
    (1-\prod_{i=(l-1)k+1}^{kl}(1-\beta_i)) + 
1-\frac{\bar\gamma_{t-1}}{\bar\gamma_{km}}
\end{align}
According to the Lemma ~\ref{Lemma:diffusion}, we have 
\begin{align}
\label{eq:SubGDiff_sampling_app}
    R^{t-1} &= \frac{1}{\mu_1}\left(R^t - \frac{\sigma_1^2}{\sqrt{\mu_1^2\sigma_2^2 + \sigma_1^2}}\epsilon_{\theta}(R^t,t)\right) +  \frac{\sigma_1 \sigma_2 }{\sqrt{\mu_1^2 \sigma_2^2 + \sigma_1^2}} z   \\
    &=\frac{1}{\sqrt{ 1-s_{km+1}\beta_t}}(R^t - \\ &\frac{s_{km+1}\beta_t}{\sqrt{(1-s_{km+1}\beta_t)(\frac{\bar\gamma_{t-1}}{\bar\gamma_{km}} p^2 \sum_{l=1}^{m}
    \frac{\bar\alpha_{m}}{\bar\alpha_{l}} 
    (1-\prod_{i=(l-1)k+1}^{kl}(1-\beta_i))+ 
1-\frac{\bar\gamma_{t-1}}{\bar\gamma_{km}}) + s_{km+1}\beta_t}} \epsilon_{\theta}(R^t,t) ) \\
    &+ \frac{\sqrt{s_{km+1}\beta_t} \sqrt{\frac{\bar\gamma_{t-1}}{\bar\gamma_{km}} p^2 \sum_{l=1}^{m}
    \frac{\bar\alpha_{m}}{\bar\alpha_{l}} 
    (1-\prod_{i=(l-1)k+1}^{kl}(1-\beta_i)) + 
1-\frac{\bar\gamma_{t-1}}{\bar\gamma_{km}}}}{\sqrt{(1-s_{km+1}\beta_t)(\frac{\bar\gamma_{t-1}}{\bar\gamma_{km}} p^2 \sum_{l=1}^{m}
    \frac{\bar\alpha_{m}}{\bar\alpha_{l}} 
    (1-\prod_{i=(l-1)k+1}^{kl}(1-\beta_i))+ 
1-\frac{\bar\gamma_{t-1}}{\bar\gamma_{km}}) + s_{km+1}\beta_t}} z
\end{align}
\begin{figure*}
    \centering
    \includegraphics[width=0.94\linewidth]{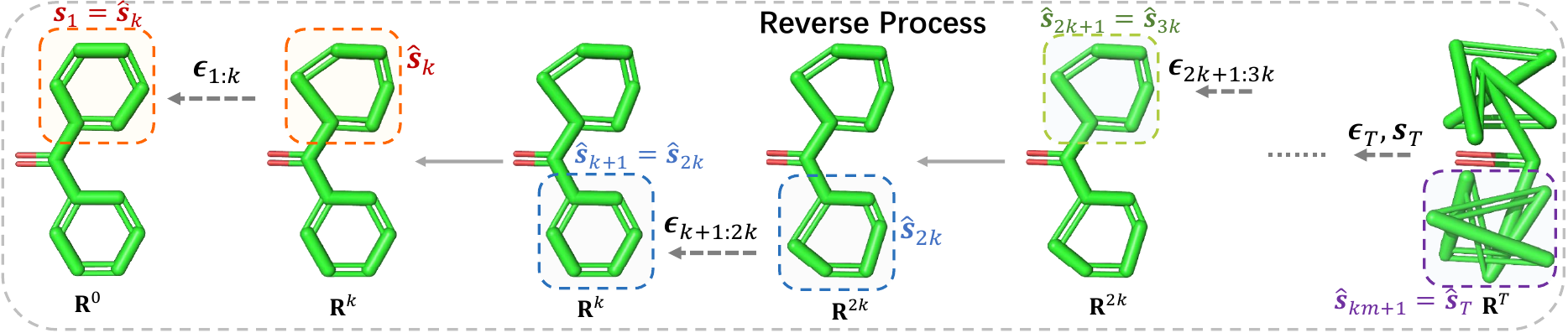}
     \caption{The reverse process of \method. The mask vector $\s$ is the same in the interval $[ki,min(ki+k,T)]$, $i=0,...,m$}
    \label{fig:sugdiff_reverse}
\end{figure*}
The schematic can see \autoref{fig:sugdiff_reverse}.

\begin{figure}[t]
\begin{algorithm}[H]
\setstretch{1.1}
\caption{Training \method } \label{alg:trainingSubGDiff_full}
\KwIn{A molecular graph $G_{3D}$, $k$ for same mask diffusion, the }
Sample $t \sim \mathcal U(1, ..., T)$ , $\epsilon \sim \mcal N(\mbf 0, \mbf I)$ \\
Sample $\mbf s^t \in p_{s_t}(\mcal S)$
\Comment*[f]
{Sample a  masked vector (subgraph node-set)} \\
$\R^t \gets  q(\R^t| \R^{0})$ \Comment*[f]{\autoref{eq:Rt_R0}} \\
$\mcal L_1=\mathrm{BCE}(\s_t, s_\vartheta(\mcal G, \R^t,t)$ \Comment*[f]{Mask prediction loss}\\
$\mcal L_2 = \|\text{diag}(\s_t) (\epsilon-\epsilon_{\theta }(\mcal G, \R^t, t))\|^2 $  \Comment*[f]{Denoising loss} \\
$\operatorname{optimizer.step}(\lambda\mcal L_1 + \mcal L_2)$ \Comment*[f]{Optimize parameters $\theta,\vartheta$}\\

\end{algorithm}\vspace{-0.10cm}

\begin{algorithm}[H]
\setstretch{1.15}
\caption{Sampling from \method }\label{alg:SamplingSubGDiff_full}
Sample $\R^T  \sim \mcal N(\mbf 0, \mbf I)$ \Comment*[f]{Random noise initialization}\\
\For{t = T \KwTo 1}{
    $\mbf z  \sim \mcal N(\mbf 0, \mbf I)$ if $t>1$, else $\mbf z =\mbf 0$ \Comment*[f]{Random noise} \\
    \textbf{If} $t\%k==0$  or $t==T$: $\hat \s \gets s_\vartheta(\mcal G,\R^{t},t)$\Comment*[f]{Mask vecter prediction} \\
    $ \hat\epsilon \gets \epsilon_\theta (\mcal G,\R^{t},t)$ \Comment*[f]{Posterior}\\
    $\R^{t-1} \gets$ \autoref{eq:SubGDiff_sampling} \Comment*[f]{sampling} 
}
\Return $\R^0$
\end{algorithm}
\end{figure}

\section{Experiment details and more results}\label{sec:app_addExperiment}
The source code would be available at \href{https://anonymous.4open.science/r/SubGDiff}{github}.
All models are trained with SGD using the ADAM optimizer.

\begin{table}[htbp]
    \centering
    \caption{Additional hyperparameters of our \method.}
    \label{app:tab:hyperparameters}
        \begin{tabular}{c cccccccc}
        \toprule
        Task & $\beta_1$ & $\beta_T$ & $\beta$ scheduler & $T$ & k (k-same mask) & $\tau$ & Batch Size & Train Iter.\\ 
        \midrule
        QM9 & 1e-7 & 2e-3 & $\operatorname{sigmoid}$ & 5000 & 250 & 10\AA & 64 & 2M \\
        Drugs & 1e-7 & 2e-3 & $\operatorname{sigmoid}$ & 5000 & 250& 10\AA & 32 & 6M \\
        \bottomrule
        \end{tabular}
\end{table}

\begin{table}[htbp]
    \centering
    \caption{Additional hyperparameters of our \method with different timesteps.}
    \label{app:tab:steps_hyperparameters}
        \begin{tabular}{c cccccccc}
        \toprule
        Task & $\beta_1$ & $\beta_T$ & $\beta$ scheduler & $T$ & k (k-same mask)& $\tau$ & Batch Size & Train Iter.\\ 
        \midrule
        500-step QM9 & 1e-7 & 2e-2 & $\operatorname{sigmoid}$ &  500& 25&10\AA & 64 & 2M \\
                200-step QM9 & 1e-7 & 5e-2 & $\operatorname{sigmoid}$ & 200 &10 & 10\AA & 64 & 2M \\
         500-step Drugs & 1e-7 & 2e-2 & $\operatorname{sigmoid}$ & 500 & 25& 10\AA & 32 & 4M \\
          1000-step Drugs & 1e-7 & 9e-3 & $\operatorname{sigmoid}$ & 500 & 50& 10\AA & 32 & 4M \\
        \bottomrule
        \end{tabular}
\end{table}

\begin{table*}[h]
     \caption{{Results on the \textbf{GEOM-Drugs} dataset under different diffusion timesteps. DDPM~\citep{ho2020denoising} is the sampling method used in GeoDiff and Langevin dynamics~\citep{song2019generative} is a typical sampling method used in DPM. Our proposed sampling method (Algorithm \ref{alg:SamplingSubGDiff}) can be viewed as a DDPM variant. \up/\down denotes  \method outperforms/underperforms  \GeoDiff.
    The threshold $\delta=1.25\si{\angstrom}$}.}
    \label{tab:drugs}
    \centering
    \begin{tabular}{lcl|llll}
    \toprule[1.0pt]
     & & & \multicolumn{2}{c}{\shortstack[c]{COV-R (\%) $\uparrow$}} & \multicolumn{2}{c}{\shortstack[c]{MAT-R (\si{\angstrom}) $\downarrow$ }} \\
     Models & Timesteps& Sampling method &  Mean & Median & Mean & Median  \\
    \midrule[0.8pt]
        \GeoDiff&   500& DDPM &  50.25   &   48.18& 1.3101& 1.2967 \\
        \method& 500&  DDPM (ours) &  {76.16}\up	& {86.43}\up	&{1.0463}\up	&{1.0264}\up \\
     \midrule[0.3pt]
         \GeoDiff&  500& LD  &     64.12    &   75.56& 1.1444 & 1.1246   \\
        \method &  500& LD (ours) &       74.30\up   &    77.87\up & 1.0003\up & 0.9905\up  \\
    \bottomrule[1.0pt]
    \end{tabular}
\end{table*}

\subsection{Mask distribution}\label{subsec:mask_distribution}
In this paper, we pre-define the mask distribution to be a discrete distribution, 
with sample space $\chi =\{G^i_{sub}\}_{i=1}^{N}$, and $p_t(\mcal S = G^i_{sub})=1/N, t>1$, where $G^i_{sub}$ is the subgraph split by the Torsional-based decomposition methods~\citep{jing2022torsional}.
The decomposition approach will cut off one torsional edge in a 3D molecule to make the molecule into two components, each of which contains at least two atoms. The two components are represented as two complementary mask vectors (i.e. $\s'+\s=\mathbf{1}$). Thus $n$ torsional edges in $G_{3D}^i$ will generate $2n$ subgraphs. Finally, for each atom $v$, the $s_{t_v} \sim Bern(0.5)$, i.e. $p=0.5$ in \method.


\subsection{Conformation Generation}\label{appsec:conformation_gene}

\begin{table}[ht]
     \caption{Results on \textbf{GEOM-QM9} dataset with different time steps. Langevin dynamics~\citep{song2019generative} is a typical sampling method used in DPM. \up denotes  \method outperforms  \GeoDiff.
    The threshold $\delta=0.5\si{\angstrom}$.}
    \label{tab:qm9_ld}
    \centering
    \resizebox{1.0\textwidth}{!}{
    \begin{tabular}{llc|llll|llll}
    \toprule[1.0pt]
     & & & \multicolumn{2}{c}{\shortstack[c]{COV-R (\%) $\uparrow$}} & \multicolumn{2}{c|}{\shortstack[c]{MAT-R (\si{\angstrom}) $\downarrow$ }}  & \multicolumn{2}{c}{\shortstack[c]{COV-P (\%) $\uparrow$}}  & \multicolumn{2}{c}{\shortstack[c]{MAT-P (\si{\angstrom}) $\downarrow$ }} \\
    Steps& Sampling method &  Models & Mean & Median & Mean & Median & Mean & Median & Mean & Median \\
    \midrule[0.8pt]
       500&Langevin dynamics &\GeoDiff & 87.80	&93.66&	0.3179	&0.3216	&46.25&	45.02&	0.6173	&0.5112 \\
       500&Langevin dynamics & \method &  91.40\up    &  95.39\up  &  0.2543\up  & 0.2601\up & 51.71\up  &   48.50\up& 0.5035\up & 0.4734\up \\
\midrule[0.5pt]
200& Langevin dynamics & \GeoDiff&       86.60   &   93.09 & 0.3532 & 0.3574 & 42.98     &  42.60 & 0.5563 & 0.5367 \\
200& Langevin dynamics & \method &       90.36\up   &   95.93\up &  0.3064\up & 0.3098\up & 48.56\up  &    46.46\up  & 0.5540\up & 0.5082\up \\
    \bottomrule[1.0pt]
    \end{tabular}
    }
    \vspace{-0.3cm}. 
\end{table}

\textbf{Evaluation metrics for conformation generation.}
To compare the generated and ground truth conformer ensembles, we employ the same evaluation metrics as in a prior study ~\citep{ganea2021geomol}: Average Minimum RMSD (AMR) and Coverage. These metrics enable us to assess the quality of the generated conformers from two perspectives: Recall (R) and Precision (P). Recall measures the extent to which the generated ensemble covers the ground-truth ensemble, while Precision evaluates the accuracy of the generated conformers.

The four metrics built upon root-mean-square deviation (RMSD), which is defined as the normalized Frobenius norm of two atomic coordinates matrices, after alignment by Kabsch algorithm~\citep{kabsch1976solution}. Formally, let $S_g$ and $S_r$ denote the sets of generated and reference conformers respectively, then the \textbf{Cov}erage and \textbf{Mat}ching metrics~\citep{xu2021cgcf} 
can be defined as:
\begin{align}
    \operatorname{COV-R}(S_g, S_r) &= \frac{1}{\vert S_r \vert} \Big\vert
    \Big\{\gC\in S_r \vert \operatorname{RMSD}(\gC, \hat{\gC}) \le \delta,  \hat{\gC} \in S_g\Big\}
    \Big\vert, \\
    \operatorname{MAT-R}(S_g, S_r) &= \frac{1}{\vert S_r \vert}
    \sum\limits_{\gC \in S_r}
    \min\limits_{\hat{\gC} \in S_g} \operatorname{RMSD}(\gC, \hat{\gC}),
\end{align}
where $\delta$ is a threshold. The other two metrics COV-P and MAT-P can be defined similarly but with the generated sets $S_g$ and reference sets $S_r$ exchanged.
In practice, $S_g$ is set as twice of the size of $S_r$ for each molecule.

 \textbf{Settings}. For \GeoDiff~\citep{xu2022geodiff} with 5000 steps, we use the checkpoints released in public \href{https://github.com/MinkaiXu/GeoDiff}{GitHub} to reproduce the results. For 200 and 500 steps, we retrain it and do the DDPM sampling.  

\paragraph{Comparison with \GeoDiff using Langevin Dynamics sampling method.}
In order to verify that our proposed diffusion process can bring benefits to other sampling methods, we conduct the experiments to compare our proposed diffusion model with \GeoDiff by adopting a typical sampling method Langevin dynamics (LD sampling)\citep{song2019generative} :
\begin{align}
    \R^{t-1} = \R^t + \alpha_t \epsilon_\theta(\mcal G, \R^t,t) + \sqrt{2\alpha_t} \mbf z_{t-1}
\end{align}
where $\mbf z_t \sim \mcal N (\mbf 0, \mbf I) $ and $h \sigma_t^2$. $h$ is the hyper-parameter referring to step size and $\sigma_t$ is the
noise schedule in the forward process. 
We use various time-step to evaluate the generalization and robustness of the proposed method, and the results shown in Table ~\ref{tab:qm9_ld} indicate that our method significantly outperforms \GeoDiff, especially when the time-step is relatively small (200,500), which implies that our training method can effectively improve the efficiency of denoising.

\begin{table}[ht]
     \caption{Results on \textbf{GEOM-QM9} dataset.
    The threshold $\delta=0.5\si{\angstrom}$.}
    \label{tab:qm9_SOTA}
    \centering
    \resizebox{0.7\textwidth}{!}{
    \begin{tabular}{l|cccc|cccc}
    \toprule[1.0pt]
      & \multicolumn{2}{c}{\shortstack[c]{COV-R (\%) $\uparrow$}} & \multicolumn{2}{c|}{\shortstack[c]{MAT-R (\si{\angstrom}) $\downarrow$ }}  & \multicolumn{2}{c}{\shortstack[c]{COV-P (\%) $\uparrow$}}  & \multicolumn{2}{c}{\shortstack[c]{MAT-P (\si{\angstrom}) $\downarrow$ }} \\
    Models & Mean & Median & Mean & Median & Mean & Median & Mean & Median \\
    \midrule[0.8pt]
    \CVGAE  & 0.09 & 0.00 & 1.6713 & 1.6088 & - & - & - & - \\ 
    \GraphDG &  73.33 & 84.21 & 0.4245 & 0.3973 & 43.90 & 35.33 & 0.5809 & 0.5823 \\ 
    \CGCF &  78.05 & 82.48 & 0.4219 & 0.3900 & 36.49 & 33.57 & 0.6615 & 0.6427 \\
    \ConfVAE &  77.84 & 88.20 & 0.4154 & 0.3739 & 38.02 & 34.67 & 0.6215 & 0.6091 \\
    \GeoMol &  71.26 & 72.00 & 0.3731 & 0.3731 & - & - & - & - \\ 
    \ConfGF &  \underline{88.49} &  \underline{94.31} &\underline{ 0.2673} & \underline{0.2685} & 46.43 & 43.41 & \bf 0.5224 & 0.5124 \\
      \GeoDiff&   80.36 & 83.82 
 &   {0.2820} &  {0.2799}   & \bf 53.66 & \bf 50.85  & 0.6673   & \bf 0.4214 \\
      \midrule[0.3pt]

        \bf \method &  \textbf{90.91} & \textbf{95.59} & \bf{0.2460} & \bf{0.2351} &  \underline{50.16} &  \underline{48.01} & \underline{0.6114} &  \underline{0.4791}\\
    \bottomrule[1.0pt]
    \end{tabular}
    }
    \vspace{-0.3cm}. 
\end{table}


\paragraph{Comparison with SOTAs.} \textbf{i) Baselines:} We compare \method with $7$ state-of-the-art baselines:
\CVGAE~\citep{mansimov2019molecular}, \GraphDG~\citep{simm2020GraphDG}, \CGCF~\citep{xu2021cgcf}, \ConfVAE~\citep{xu2021end}, \ConfGF~\citep{shi2021learning} and \GeoDiff~\citep{xu2022geodiff}. 
For the above baselines, we reuse the experimental results reported by \cite{xu2022geodiff}.
For \GeoDiff~\citep{xu2022geodiff}, we use the checkpoints released in public \href{https://github.com/MinkaiXu/GeoDiff}{GitHub} to reproduce the results.
\textbf{ii)Results:}
The results on the GEOM-QM9 dataset are reported in Table \ref{tab:qm9_SOTA}. 
From the results, we get the following observation: \method significantly outperforms the baselines on COV-R, indicating the \method tends to explore more possible conformations. This implicitly demonstrates the subgraph will help fine-tune the generated conformation to be a potential conformation.

\paragraph{Model Architecture.}
We adopt the graph field network (GFN) from \cite{xu2022geodiff} as the GNN encoder for extracting the 3D molecular information.
In the $l$-th layer, the GFN receives node embeddings $\mathbf{h}^l \in \mathbb{R}^{n \times b}$ (where $b$ represents the feature dimension) and corresponding coordinate embeddings $\mathbf{x}^l \in \mathbb{R}^{n \times 3}$ as input. It then produces the output $\mathbf{h}^{l+1}$ and $\mathbf{x}^{l+1}$ according to the following process:
\begin{align}
    & \mathbf{m}_{ij}^l =\Phi^l_{m}\left(\mathbf{h}_{i}^{l}, \mathbf{h}_{j}^{l},\|\mathbf{x}_{i}^{l}-\mathbf{x}_{j}^{l}\|^{2}, e_{ij}; \theta_m \right) \label{eq:nfl-message} \\
    & \mathbf{h}_{i}^{l+1} =\Phi^l_{h}\Big(\mathbf{h}_{i}^l, \sum_{j \in \mathcal{N}(i)} \mathbf{m}_{ij}^l; \theta_h \Big) \label{eq:nfl-node} \\
    & \mathbf{x}_{i}^{l+1} = \sum_{j \in \mathcal{N}(i)}\frac{1}{d_{ij}}\left(R_{i} - R_{j}\right) \Phi^l_{x}\left(\mathbf{m}_{ij}^l ; \theta_x \right) \label{eq:nfl-tensor}  
\end{align}
where $\Phi$ are implemented as feed-forward networks and $d_{ij}$ denotes interatomic distances. The initial embedding $\mathbf{h}^0$ is composed of atom embedding and time step embedding while $\mathbf{x}^0$ represents atomic coordinates. $\gN(i)$ is the neighborhood of $i^{th}$ node, consisting of connected atoms and other ones within a radius threshold $\tau$, helping the model capture long-range interactions explicitly and support disconnected molecular graphs. 

\begin{figure*}
    \centering
    \includegraphics[width=0.999\linewidth]{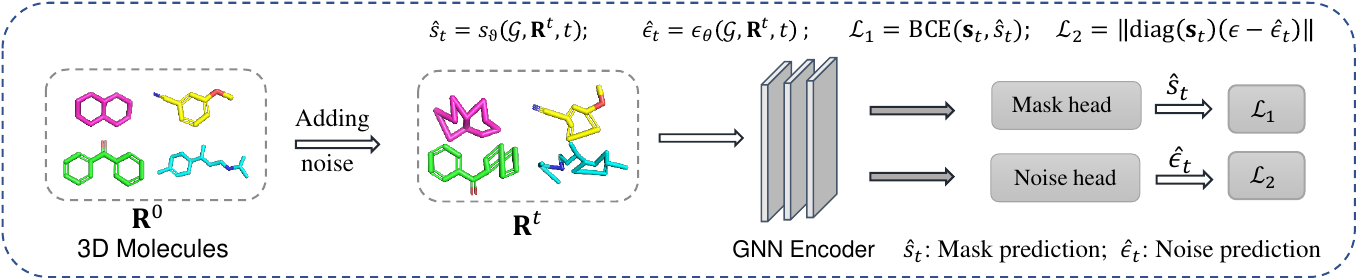}
     \caption{The model architecture for denoising \method.}
    \label{fig:Training_framework}
        \vspace{-4mm}
\end{figure*}

Eventually, the Gaussian noise and mask can be predicted as follows (C.f. \autoref{fig:Training_framework}):
\begin{align}
& \hat{\epsilon}_i = \mathbf{x}_i^L \\
    & \hat{s}_i = \text{MLP}(\mathbf{h}_i^L) 
\end{align}
where $\hat{\epsilon}_i$ is equivalent and $\hat{s}_i$ is invariant.

\subsection{Domain generelizaion}
The results of  Training on QM9 (small molecular with up to 9 heavy atoms) and testing on Drugs (medium-sized organic compounds) can be found in table \ref{tab:Drugs_DG}.
\begin{table}[ht]
    \caption{Results on the \textbf{GEOM-Drugs} dataset. The threshold $\delta=1.25\si{\angstrom}$}
    \label{tab:Drugs_DG}
    \vspace{-0.4cm}
    \centering
    \scalebox{0.68}{
    \begin{tabular}{lc|cccc}
    \toprule[1.0pt]
     & Train& \multicolumn{2}{c}{\shortstack[c]{COV-R (\%) $\uparrow$}} & \multicolumn{2}{c}{\shortstack[c]{MAT-R (\si{\angstrom}) $\downarrow$ }}   \\
    Models &  data & Mean & Median & Mean & Median  \\
    \midrule[0.8pt]
    \CVGAE & Drugs & 0.00 & 0.00 & 3.0702 & 2.9937  \\ 
    \GraphDG & Drugs &  8.27 & 0.00 & 1.9722& 1.9845 \\ 

 \GeoDiff&  QM9& 7.99
 & 0.00 &  2.7704 &  2.3297   \\

      \midrule[0.3pt]
    \bf \method &  QM9 &\textbf{24.01} & \textbf{9.93} & \bf{1.6128} & \bf{1.5819} \\
    \bottomrule[1.0pt]
    \end{tabular}
    }
\end{table}

\subsection{Self-supervised learning} \label{appsec:SSL}

\subsubsection{Model architecture}
 We use the pretraining framework MoleculeSDE proposed by \cite{liu2023group_molecularsde} and extend our \method to multi-modality pertaining. The two key components of MoleculeSDE are two SDEs(stochastic differential equations ~\citet{song2020score}): an
SDE from 2D topology to 3D conformation ($\twoD \to \threeD$)
and an SDE from 3D conformation to 2D topology ($\threeD \to \twoD$). In practice, these two SDEs can be replaced by discrete diffusion models. In this paper, we use the proposed \method to replace the SDEs.

\textbf{2D topological molecular graph.}
A topological molecular graph is denoted as $g_{\twoD{}} = \mcal G(\mcal V, \E, \X)$, where $\X$ is the atom attribute matrix and $\X$ is the bond attribute matrix. The 2D graph representation with graph neural network (GNN) is:
\begin{equation} \label{eq:2d_gnn}
\vx \triangleq \mH_{\twoD{}} = \text{GIN}(g_{\twoD{}}) = \text{GIN}(\X, \X),
\end{equation}
where GIN is the a powerful 2D graph neural network~\citep{xu2018powerful} and $\mH_{\twoD} = [h_{\twoD{}}^0, h_{\twoD{}}^1, \hdots]$, where $h_{\twoD{}}^i$ is the $i$-th node representation. 

\textbf{3D conformational molecular graph.} 
The molecular conformation is denoted as $g_{\threeD{}} := G_{3D}(\mcal G, \R)$.
The conformational representations are obtained by a 3D GNN SchNet~\citep{schutt2017schnet}:
\begin{equation} \label{eq:3d_gnn}
\vy \triangleq  \mH_{\threeD{}} = \text{SchNet}(g_{\threeD{}}) = \text{SchNet}(\gG, \R),
\end{equation}
where  $\mH_{\threeD{}} = [h_{\threeD{}}^0, h_{\threeD{}}^1, \hdots]$, and $h_{\threeD{}}^i$ is the $i$-th node representation.

\paragraph{An SE(3)-Equivariant Conformation Generation} \label{sec:generative_ssl_2D_to_3D}
The first objective is the conditional generation from topology to conformation, $p(\vy|\vx)$, implemented as \method.
The denoising network we adopt is the SE(3)-equivariance network ($S_\theta^{\twoD{} \rightarrow \threeD{}}$) used in MoleculeSDE. The details of the network architecture refer to ~\cite{liu2023group_molecularsde}.

Therefore, the training objective from 2D topology graph to 3D confirmation is:
\begin{equation}
\begin{aligned}
\mathcal{L}_{\twoD{} \rightarrow \threeD{}}
& = \mathbb{E}_{\vx,\R, t, \s_t} \mathbb{E}_{\R_t | \R} \\
& \Big[ \Big\| \text{diag}(\s_t)(\mbf{\epsilon}- S_\theta^{\twoD{} \rightarrow \threeD{}} (\vx, \R_t, t))\Big\|_2^2 + \mathrm{BCE}(\s_t,s^{\twoD{} \rightarrow \threeD{}}_\vartheta(\vx, \R_t, t)) \Big],
\end{aligned}
\end{equation}
where $s^{\twoD{} \rightarrow \threeD{}}_\vartheta(\vx, \R_t, t)$ gets the invariant feature from $S_\theta$ and introduces a mask head (MLP)  to read out the mask prediction.

\textbf{{An SE(3)-Invariant Topology Generation}.} \label{sec:generative_ssl_3D_to_2D}
The second objective is to reconstruct the 2D topology from 3D conformation, i.e., $p(\vx|\vy)$. We also use the SE(3)-invariant score network $S_\theta^{\threeD{} \rightarrow \twoD{}}$  proposed by MoleculeSDE. The details of the network architecture refer to ~\cite{liu2023group_molecularsde}.
For modeling $S_\theta^{\threeD{} \rightarrow \twoD{}}$, it needs to satisfy the SE(3)-invariance symmetry property. The inputs are 3D conformational representation $\vy$, the noised 2D information $\vx_t$ at time $t$, and time $t$. The output of $S_\theta^{\threeD{} \rightarrow \twoD{}}$ is the Gaussian noise, as $(\epsilon^{\X}, \epsilon^{\E})$. The diffused 2D information contains two parts: $\vx_t = (\X_t, \E_t)$. For node feature $\X$, the training objective is
\begin{align}
   \scalemath{0.82}{\mathcal{L}_{\threeD{} \rightarrow \twoD{}}^{\X}}
&= \scalemath{0.82}{\mathbb{E}_{\X,\vy} \mathbb{E}_{t,\s_t}  \scalemath{0.88}{\mathbb{E}_{\X_t | \X} }}\\
&  \scalemath{0.82}{\Big[ \Big\| \text{diag}(\s_t)(\epsilon- S_\theta^{\threeD{} \rightarrow \twoD{}} (\vy, \X_t, t))\Big\|_2^2 + \mathrm{BCE}(\s_t,s^{\threeD{} \rightarrow \twoD{}}_\vartheta(\vy, \X_t, t)) \Big] }.
\end{align}
 For edge feature $\E$, we define a mask matrix $\rmS$ from mask vector $\s$: $\mathbf S_{ij}=1$ if $\s_i=1$ or $\s_j=1$, otherwise, $\mathbf S_{ij}=0$. Eventually, the ojective can be written as:
\begin{align}
     \scalemath{0.82}{\mathcal{L}^{\E}_{\threeD{} \rightarrow \twoD{}}}
& = \scalemath{0.82}{\mathbb{E}_{\E,\vy} \mathbb{E}_{t,\s_t} \mathbb{E}_{\E_t | \E} }\\
& \scalemath{0.82}{ \Big[ \Big\| \rmS_t \odot (\epsilon- S_\theta^{\threeD{} \rightarrow \twoD{}} (\vy, \E_t, t))\Big\|_2^2 + \mathrm{BCE}(\s_t,s_\vartheta^{\threeD{} \rightarrow \twoD{}}(\vy, \E_t, t)) \Big]},
\end{align}
Then the score network $S_\theta^{\threeD{} \rightarrow \twoD{}}$ is also decomposed into two parts for the atoms and bonds: $S_\theta^{\X_t}(\vx_t)$ and $S_\theta^{\E_t}(\vx_t)$. Similarly, the mask predictor $s_\vartheta^{\threeD{} \rightarrow \twoD{}}$ is also decomposed into two parts for the atoms and bonds: $s_\vartheta^{\X_t}(\vx_t)$ and $s_\vartheta^{\E_t}(\vx_t)$. 

Similar to the topology to conformation generation procedure, 
the $s^{\threeD{} \rightarrow \twoD{}}_\vartheta(\vx, \R_t, t)$ gets the invariant feature from $S^{\threeD{} \rightarrow \twoD{}}_\theta$ and introduces a mask head (MLP)  to read out the mask prediction.

\textbf{Learning.} Following MoleculeSDE, we incorporate a contrastive loss called EBM-NCE~\cite{liu2022pretraining}. EBM-NCE provides an alternative approach to estimate the mutual information $I(X;Y)$ and is anticipated to complement the generative self-supervised learning (SSL) method. As a result, the ultimate objective is:
\begin{equation}
\mathcal{L}_{\text{overall}} = \alpha_1 \mathcal{L}_{\text{Contrastive}} + \alpha_2 \mathcal{L}_{\twoD{} \rightarrow \threeD{}} + \alpha_3 (\mathcal{L}^{\X}_{\threeD{} \rightarrow \twoD{}}+\mathcal{L}^{\E}_{\threeD{} \rightarrow \twoD{}}),   
\end{equation}
where $\alpha_1, \alpha_2, \alpha_3$ are three coefficient hyperparameters.

\subsubsection{Dataset and settings}\label{appsubsec:setting}
\textbf{Dataset.}
For pretraining, following MoleculeSDE, we use PCQM4Mv2~\citep{hu2020ogb}. It's a sub-dataset of PubChemQC~\citep{nakata2017pubchemqc} with 3.4 million molecules with both the topological graph and geometric conformations. 

For finetuning, in addition to QM9~\citep{ramakrishnan2014quantum}, we also include MD17. To be specific,
MD17 comprises eight molecular dynamics simulations focused on small organic molecules. These datasets were initially presented by \citet{chmiela2017machine} for the development of energy-conserving force fields using GDML.
 Each dataset features the trajectory of an individual molecule, encompassing a broad spectrum of conformations. The objective is to predict energies and forces for each trajectory by employing a single model.

\paragraph{Baselines for 3D property prediction}
We begin by incorporating three coordinate-MI-unaware SSL methods: (1) Type Prediction, which aims to predict the atom type of masked atoms; (2) Angle Prediction, which focuses on predicting the angle among triplet atoms, specifically the bond angle prediction; (3) 3D InfoGraph, which adopts the contrastive learning paradigm by considering the node-graph pair from the same molecule geometry as positive and negative otherwise. Next, in accordance with the work of \citep{liu2023molecular_geossl}, we include two contrastive baselines: (4) GeoSSL-InfoNCE \citep{oord2018representation} and (5) GeoSSL-EBM-NCE \citep{liu2022pretraining}. We also incorporate a generative SSL baseline named (6) GeoSSL-RR (RR for Representation Reconstruction). The above baselines are pre-trained on a subset of 1M molecules with 3D geometries from Molecule3D~\citep{xu2021molecule3d} and we reuse the results reported by ~\cite{liu2023molecular_geossl} with SchNet as backbone.

\textbf{Baselines for 2D topology pretraining.}
We pick up the most promising ones as follows. AttrMask~\citep{hu2020strategies,liu2019n}, ContexPred~\citep{hu2020strategies},  InfoGraph~\citep{sun2019infograph}, and MolCLR~\citep{wang2022molecular}.

\textbf{Baselines for 2D and 3D multi-modality pretraining.}
 We include MoleculeSDE\citep{liu2023group_molecularsde}(Variance Exploding (VE) and Variance
Preserving (VP)) as a crucial baseline to verify the effectiveness of our methods due to the same pertaining framework. We reproduce the results from the released \href{https://github.com/chao1224/MoleculeSDE}{Code}.

\paragraph{Compared with \GeoDiff.}
We directly reuse the pre-trained model of the molecular conformation generation in sec. \ref{subsec:Confor_gene}
for fine-tuning, to compare our method with \GeoDiff from naive denoising pretraining perspective~\citep{zaidi2022pre_deepMind}. The results are shown in Table 
\ref{tab:schnet_QM9_geodiff}.

\begin{table}[htb]
\setlength{\tabcolsep}{5pt}
\vspace{-0.2cm}
\fontsize{9}{9}\selectfont
\caption{
\small{
Results on 12 quantum mechanics prediction tasks from QM9. We take 110K for training, 10K for validation, and 11K for testing. The evaluation is mean absolute error~(MAE), and the best and the second best results are marked in bold and underlined, respectively.  The backbone is \textbf{SchNet}.
}
}
\label{tab:schnet_QM9_geodiff}
\vspace{-2ex}
\begin{adjustbox}{max width=\textwidth}
\begin{tabular}{l c c c c c c c c c c c c}
\toprule
Pretraining & Alpha $\downarrow$ & Gap $\downarrow$ & HOMO$\downarrow$ & LUMO $\downarrow$ & Mu $\downarrow$ & Cv $\downarrow$ & G298 $\downarrow$ & H298 $\downarrow$ & R2 $\downarrow$ & U298 $\downarrow$ & U0 $\downarrow$ & Zpve $\downarrow$\\
\midrule
\GeoDiff & 0.078 & 51.84 & 30.88 & 28.29 & 0.028 & 0.035 & 15.35 &11.37 &  0.132 & 15.76 & 15.24 &  1.869\\
\midrule
\bf \method (ours) & 0.076\up & {50.80}\up& 31.15\down &{26.62}\up & {0.025}\up & {0.032}\up & 14.92\up & 12.86\up  & 0.129\up & 14.74\up& 14.53\up& 1.710\up\\

\bottomrule
\end{tabular}
\end{adjustbox}
\vspace{-0.3cm}
\end{table}

\subsubsection{3D molecular property prediction Results on QM9.}\label{appsubsec:3DQM9}

By adopting the pertaining setting in Appendix \ref{appsubsec:setting}, we also take the QM9 dataset for finetuning and follow the literature~\citep{schutt2017schnet, schutt2021equivariant, liu2023group_molecularsde}, using 110K for training, 10K for validation and 11k for testing.
In addition, the QM9 dataset encompasses 12 tasks that pertain to quantum properties, which are commonly used for evaluating representation learning tasks~\citep{schutt2017schnet,liu2023molecular_geossl}. 
The experimental results can be seen in Table \ref{tab:schnet_QM9_result}. The results also suggest the superior performance of our method.

\begin{table*}[htb]
\setlength{\tabcolsep}{5pt}

\fontsize{9}{9}\selectfont
\caption{
\small{\revise{
Results on 12 quantum mechanics prediction tasks from QM9. We take 110K for training, 10K for validation, and 11K for testing. The evaluation is mean absolute error~(MAE), and the best and the second best results are marked in bold and underlined, respectively.  The backbone is \textbf{SchNet}.}
}
}
\label{tab:schnet_QM9_result}
\vspace{-2ex}
\begin{adjustbox}{max width=\textwidth}
\begin{tabular}{l c c c c c c c c c c c c}
\toprule
Pretraining & Alpha $\downarrow$ & Gap $\downarrow$ & HOMO$\downarrow$ & LUMO $\downarrow$ & Mu $\downarrow$ & Cv $\downarrow$ & G298 $\downarrow$ & H298 $\downarrow$ & R2 $\downarrow$ & U298 $\downarrow$ & U0 $\downarrow$ & Zpve $\downarrow$\\
\midrule
 Random init & 0.070 & 50.59 & 32.53 & 26.33 & 0.029 & 0.032 & 14.68 & 14.85 & 0.122 & 14.70 & 14.44 & 1.698\\
Supervised & 0.070 & 51.34 & 32.62 & 27.61 & 0.030 & 0.032 & 14.08 & 14.09 & 0.141 & 14.13 & 13.25 & 1.727\\
Type Prediction & 0.084 & 56.07 & 34.55 & 30.65 & 0.040 & 0.034 & 18.79 & 19.39 & 0.201 & 19.29 & 18.86 & 2.001\\
Distance Prediction & 0.068 & 49.34 & 31.18 & 25.52 & 0.029 & 0.032 & 13.93 & 13.59 & 0.122 & 13.64 & 13.18 & 1.676\\
Angle Prediction & 0.084 & 57.01 & 37.51 & 30.92 & 0.037 & 0.034 & 15.81 & 15.89 & 0.149 & 16.41 & 15.76 & 1.850\\
3D InfoGraph & 0.076 & 53.33 & 33.92 & 28.55 & 0.030 & 0.032 & 15.97 & 16.28 & 0.117 & 16.17 & 15.96 & 1.666\\
GeossL-RR & 0.073 & 52.57 & 34.44 & 28.41 & 0.033 & 0.038 & 15.74 & 16.11 & 0.194 & 15.58 & 14.76 & 1.804\\
GeossL-InfoNCE & 0.075 & 53.00 & 34.29 & 27.03 & 0.029 & 0.033 & 15.67 & 15.53 & 0.125 & 15.79 & 14.94 & 1.675\\
GeossL-EBM-NCE & 0.073 & 52.86 & 33.74 & 28.07 & 0.031 & 0.032 & 14.02 & 13.65 & 0.121 & 13.70 & 13.45 & 1.677\\
GeossL   & {0.066} & {48.59} & {30.83} & {25.27} & {0.028} & {0.031} & {13.06} & {12.33} & {0.117} & {12.48} & {12.06} & {1.631}\\
MoleculeSDE & 0.062 & 47.74 & 28.02 &{24.60} & { 0.028} & {0.029} & 13.25 &12.70 & 0.120 & 12.68& 12.93& 1.643\\
\midrule
\textbf{Ours} & \textbf{0.054} & \textbf{44.88} & \textbf{25.45} &\textbf{23.75} & \textbf{ 0.027} & \textbf{0.028} & \textbf{12.03} & \textbf{11.46}& \textbf{0.110} & \textbf{11.32}& \textbf{11.25}& \textbf{1.568}\\

\bottomrule
\end{tabular}
\end{adjustbox}
\vspace{-2mm}
\end{table*}


\subsection{Visualization}

We conduct an alignment analysis to validate that our method can capture chemically informative subgraphs during pretraining. Specifically, we employ t-distributed stochastic neighbor embedding (t-SNE) to represent molecules with various scaffolds visually. The purpose is to investigate whether molecules sharing the same scaffold exhibit similar representations, extracted by the pretrained molecular encoder. 
A scaffold is usually represented by a substructure of a molecule and can be regarded as the subgraph in our \method.

In our analysis, we select the nine most prevalent scaffolds from each dataset (BBBP, Sider, ClinTox, and Bace) and assign each molecule to a cluster according to its scaffold. To quantify the molecule embedding, we compute the Silhouette index of the embeddings for each dataset.

\begin{figure}[htb]
  \centering
    \subfigure{
    \includegraphics[width=\textwidth]{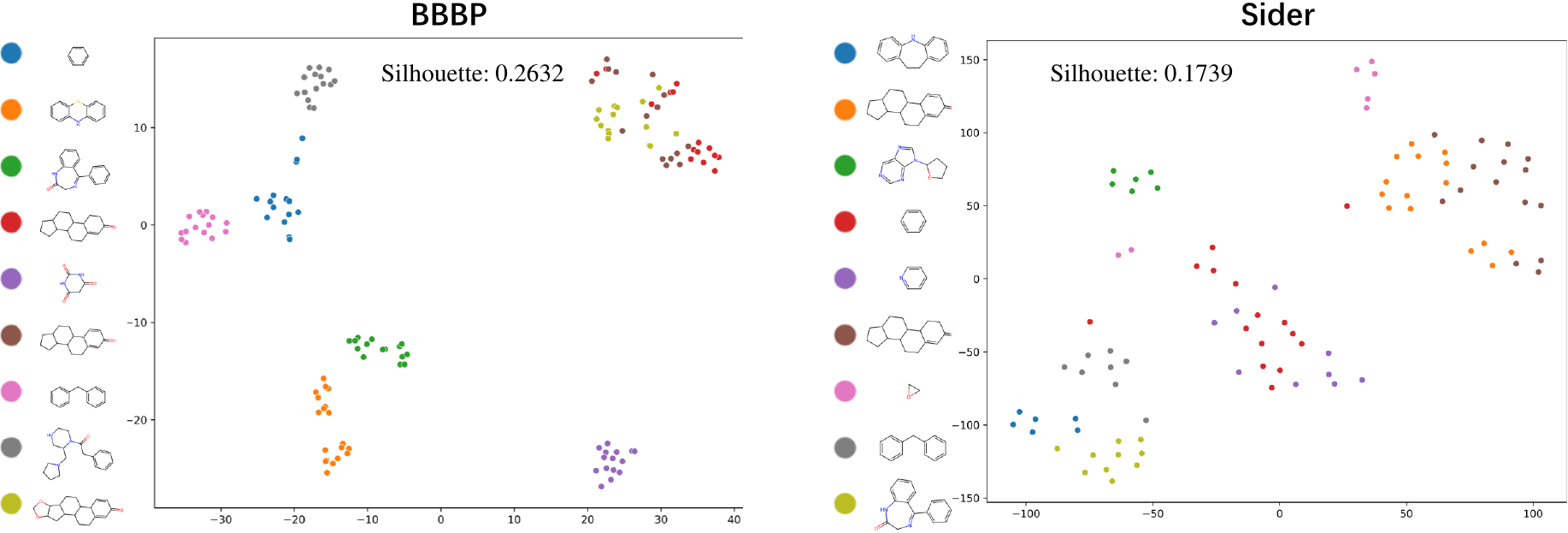}
}
  \hfill
    \subfigure{
    \includegraphics[width=\textwidth]{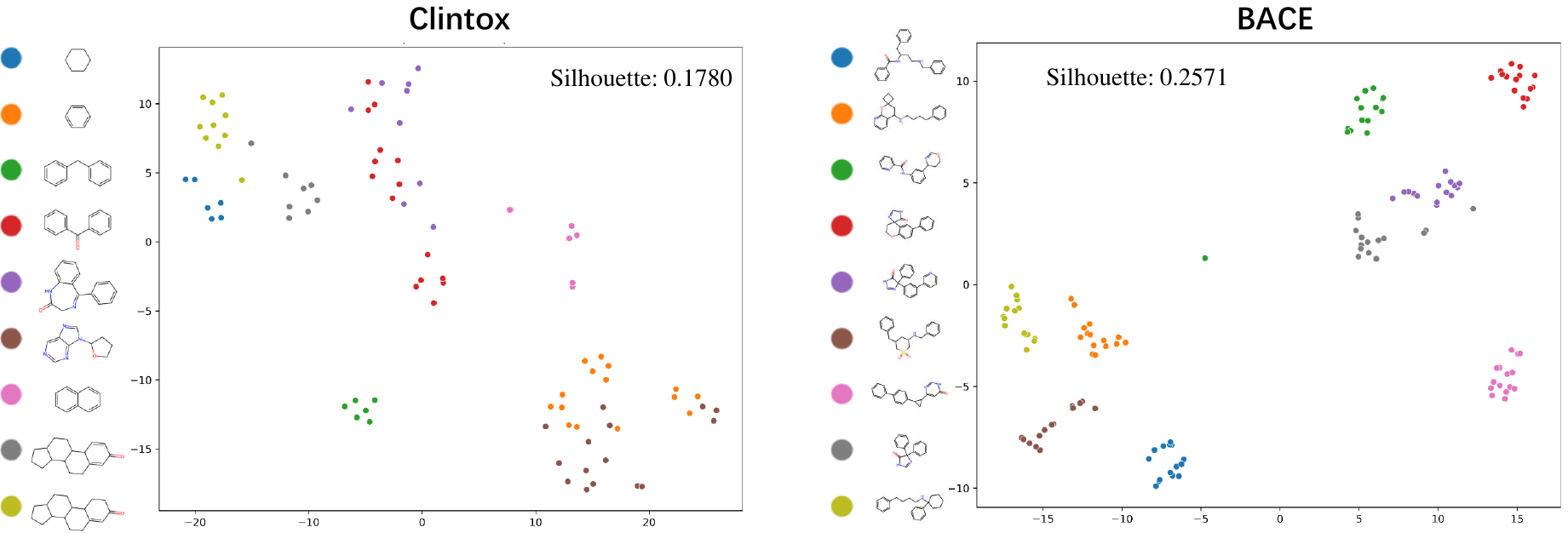}
}
  \caption{T-distributed stochastic neighbor embedding (t-SNE) visualization of the learned molecules representations, colored by the scaffolds of the molecules.}
  \label{fig:t-sne}
\end{figure}

As shown in \autoref{tab:t-sne}, \method enables the generation of more distinctive representations of molecules with different scaffolds. This implies that \method enables the denoising network (molecular encoder) to better capture the subgraph (scaffold) information. We also provide the t-SNE visualizations in \autoref{fig:t-sne}.

\begin{table}[htb]
\setlength{\tabcolsep}{5pt}
\vspace{-0.2cm}
\centering
\caption{
\small{Silhouette index (higher is better) of the molecule embeddings on Moleculenet dataset (with 2D topology only)
}
}
\label{tab:t-sne}
\vspace{-2ex}
\begin{adjustbox}{max width=\textwidth}
\begin{tabular}{l|ccccc}
        \toprule
       & BBBP $\uparrow$ & ToxCast $\uparrow$ & Sider $\uparrow$ & ClinTox $\uparrow$ & Bace $\uparrow$ \\
       \midrule
        MoleculeSDE    & 0.2344          & 0.0611             & 0.1664           & 0.1394              & 0.1860          \\
        \method(ours)  & \textbf{0.2632} & \textbf{0.0650}    & \textbf{0.1739}  & \textbf{0.1780}     & \textbf{0.2571} \\ 
\bottomrule
\end{tabular}
\end{adjustbox}
\vspace{-0.3cm}
\end{table}


\end{document}